\documentclass[10pt,twocolumn,letterpaper]{article}

\usepackage[pagenumbers]{cvpr} 

\usepackage[dvipsnames]{xcolor}

\usepackage{url}            
\usepackage{booktabs}       
\usepackage{amsfonts}       
\usepackage{nicefrac}       
\usepackage{microtype}      
\usepackage[dvipsnames]{xcolor}
\usepackage{multirow}
\usepackage{graphicx}
\usepackage{colortbl}
\usepackage{bm}
\usepackage{balance}
\usepackage{amsmath}
\usepackage{wrapfig}
\usepackage{adjustbox}
\usepackage{color}
\usepackage{bbding}
\usepackage[ruled]{algorithm2e}

\makeatletter
\renewcommand{\maketag@@@}[1]{\hbox{\m@th\normalsize\normalfont#1}}%
\makeatother

\newcommand{\cc}{\cellcolor{gray!20}}
\newcommand{\ttt}[1]{$\mathtt{#1}$}

\newcommand{\hiker}[0]{HiKER-SGG~}

\definecolor{cvprblue}{rgb}{0.21,0.49,0.74}
\usepackage[pagebackref,breaklinks,colorlinks,citecolor=cvprblue]{hyperref}

\def\paperID{7675} 
\def\confName{CVPR}
\def\confYear{2024}

\title{HiKER-SGG: Hierarchical Knowledge Enhanced Robust Scene Graph Generation}

\author{
Ce Zhang \quad Simon Stepputtis\quad Joseph Campbell\quad Katia Sycara\quad Yaqi Xie\vspace{1mm}\\
School of Computer Science, Carnegie Mellon University\\ 
{\tt\small \{cezhang, sstepput, jacampbe, katia, yaqix\}@cs.cmu.edu} 
}

\begin{document}
\maketitle
\begin{abstract}

Being able to understand visual scenes is a precursor for many downstream tasks, including autonomous driving, robotics, and other vision-based approaches. 
A common approach enabling the ability to reason over visual data is Scene Graph Generation (SGG); however, many existing approaches assume undisturbed vision, i.e., the absence of real-world corruptions such as fog, snow, smoke, 
as well as non-uniform perturbations like sun glare or water drops.
In this work, we propose a novel SGG benchmark containing procedurally generated weather corruptions and other transformations over the Visual Genome dataset. Further, we introduce a corresponding approach, \textbf{Hi}erarchical \textbf{K}nowledge \textbf{E}nhanced \textbf{R}obust \textbf{S}cene \textbf{G}raph \textbf{G}eneration (HiKER-SGG), providing a strong baseline for scene graph generation under such challenging setting. 
At its core, HiKER-SGG utilizes a hierarchical knowledge graph in order to refine its predictions from coarse initial estimates to detailed predictions.
In our extensive experiments, we show that \hiker does not only demonstrate superior performance on corrupted images in a zero-shot manner, but also outperforms current state-of-the-art methods on uncorrupted SGG tasks. Code is available at \href{https://github.com/zhangce01/HiKER-SGG}{https://github.com/zhangce01/HiKER-SGG}.
\vspace{-2pt}
\end{abstract}

\section{Introduction}


Visual scene understanding and the ability to extract information from images has made significant progress through the development of deep learning~\cite{eslami2016attend,xiao2018unified,chen2019towards}. 
Particularly, Scene Graph Generation (SGG)~\cite{chang2021comprehensive,zhang2023robust,zellers2018neural} from visual inputs is a powerful method of extracting semantic information from images, enabling many subsequent reasoning tasks~\cite{wald2020learning,wang2020sketching,desai2021learning,zhu2021hierarchical,li2022embodied}. 
However, most existing studies in this field assume access to ``clean'' images. This contrasts with real-world situations where images often have corruptions like sun glare, dust, water drops, and rain~\cite{halder2019physics,quan2019deep,tremblay2021rain,gray2023glare}.
Being exposed to and handling such corruptions is a challenging task for many systems as it is unlikely that models can be sufficiently trained to handle such domain shifts.
Inspired by the human ability to recognize objects in corrupted images using prior domain knowledge, our work leverages similar knowledge for scene graph generators. This not only enables accurate identification in corrupted images but also improves over state-of-the-art model performance on clean images.

\begin{figure}[t]
\centering
\includegraphics[width=\linewidth]{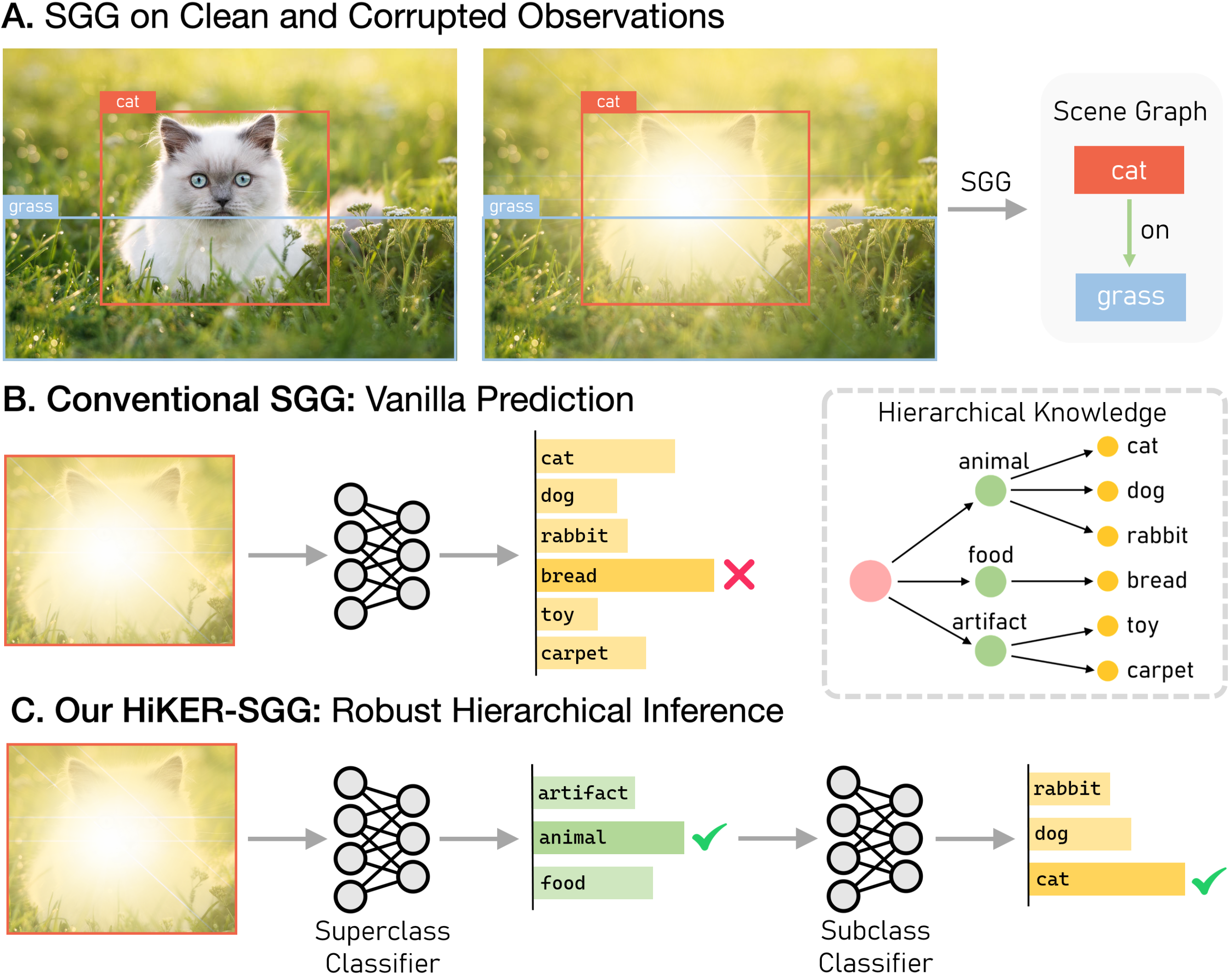}
\vspace{-18pt}
\caption{\looseness = -1
\textbf{We introduce a novel task: robust SGG in the presence of real-world corruptions}. Consider an image of a cat obscured by sun glare as an example, where conventional methods often struggle. Our \hiker leverages hierarchical knowledge to first infer the broader category of an object, for example, \ttt{animal}, before continuing to a more granular identification of an object constrained to various animals. By utilizing such an approach, we simplify the process to correctly identify it as a \ttt{cat}.}
\label{fig:intro}
\vspace{-10pt}
\end{figure}

\begin{figure*}[t]
\centering
\includegraphics[width=\linewidth]{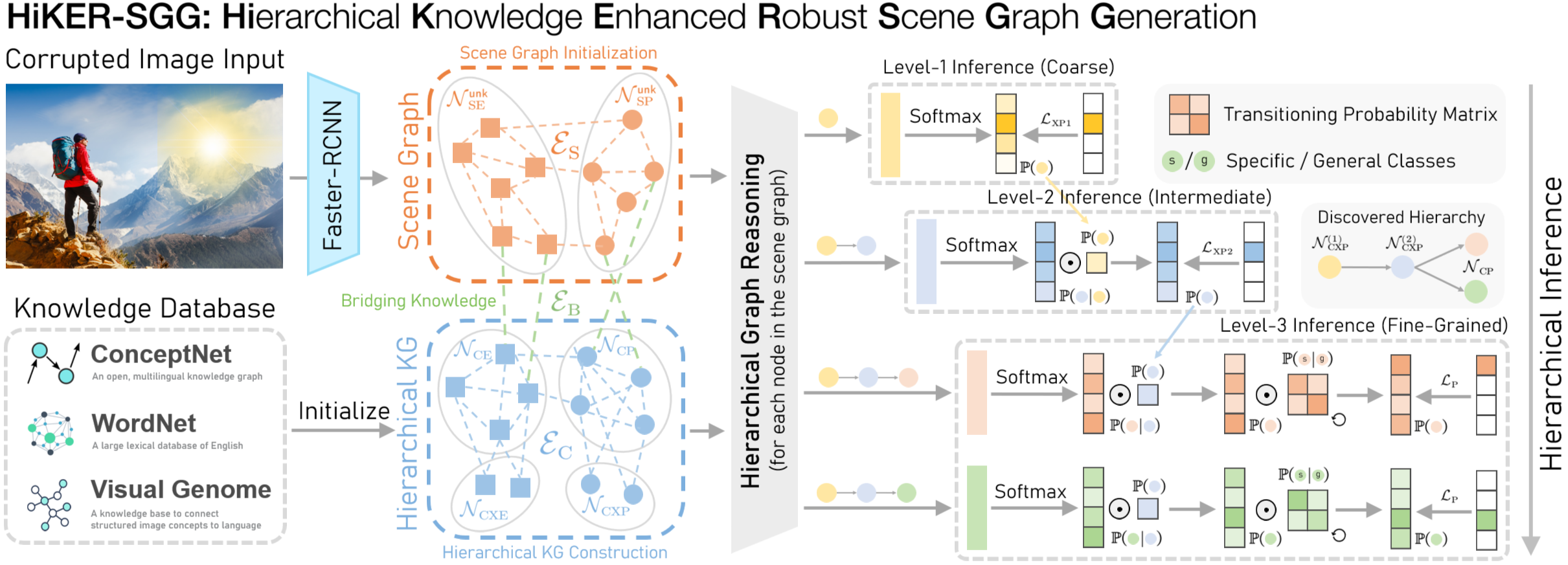}
\vspace{-18pt}
\caption{\textbf{\hiker overview}. 
Hierarchical knowledge graphs are constructed from an external knowledge base. Given an image, we first initialize the scene graph using an off-the-shelf detector, Faster-RCNN~\cite{ren2015faster}. We then create bridging connections between the hierarchical knowledge graph and the initial scene graph and perform message passing for hierarchical graph reasoning. Finally, we design a hierarchical inference process to guide the model in making step-by-step predictions explicitly.}
\label{fig:overview}
\vspace{-10pt}
\end{figure*}


In this work, we propose a novel method -- \textbf{Hi}erarchical \textbf{K}nowledge \textbf{E}nhanced \textbf{R}obust \textbf{S}cene \textbf{G}raph \textbf{G}eneration (HiKER-SGG) -- which utilizes a hierarchical approach that reasons over multiple levels of domain knowledge with increasing granularity in order to generate accurate scene graphs for both corrupted and clean images. Further, we introduce an accompanying benchmark -- Corrupted Visual Genome (VG-C) -- providing 20 procedurally generated image corruptions, resembling common transformation and various weather conditions. 
The proposed benchmark fills a crucial gap in the field of scene graph generation and offers a comprehensive evaluation platform to assess the robustness of SGG models in adverse conditions. 

Our method, HiKER-SGG, is visualized in Figure~\ref{fig:intro}: When given a previously unseen corrupted image, \hiker first identifies object candidates by utilizing a pre-trained object detector. 
For each proposed image region (\textit{e.g.}, a region surrounding a \ttt{cat}), \hiker determines the type of the object by first identifying its high-level type (\textit{e.g.}, \ttt{animal}) before proceeding to more granular predictions by selecting \ttt{cat} among the possible \ttt{animals}. 
A key benefit of our proposed hierarchical approach is that the individual classification tasks at each level of our hierarchy are simpler than learning to create detailed predictions directly. 
Through each level of our hierarchy, the search space is constrained to the children of the previously identified superclass, making \hiker a powerful method for scene graph generation, particularly in the presence of image corruptions without requiring explicit training on corrupted images. 
Making a fundamental determination whether or not the depicted object is an \ttt{animal} or an \ttt{artifact} may still be accurate despite the corruption, which allows for more accurate object classification in subsequent levels of our hierarchy. 

To evaluate the effectiveness of our proposed HiKER-SGG, we conduct comprehensive experiments on both the original clean Visual Genome (VG) dataset and our introduced VG-C benchmark. Remarkably, our proposed \hiker outperforms state-of-the-art models on clean images, and exhibits exceptional zero-shot performance in handling various types of corrupted observations. 

Our work opens new research avenues and emphasizes the need for robust vision models to handle real-world image challenges and proposes the following contributions:

\begin{itemize}
    \item We propose HiKER-SGG, a novel method for generating scene graphs through a hierarchical inference approach over structured domain knowledge, allowing it to gradually specify increasingly granular classifications through iterative sub-selection.
    \item We introduce a new synthetic VG-C benchmark for SGG, containing 20 challenging image corruptions, including simple transformations and severe weather conditions. 
    \item Extensive experiments demonstrate that \hiker outperforms current state-of-the-art methods on SGG tasks, while simultaneously providing a strong zero-shot baseline for generating scene graphs from corrupted images. 
\end{itemize}

\section{Related Work}

\textbf{Scene Graph Generation.}
Scene graph generation has emerged as a key area of focus in computer vision research, with the goal of offering a structured depiction of an image through the identification of objects and their intricate relations~\cite{chang2021comprehensive,tian2021mask}. Furthermore, numerous studies illustrate that scene graphs can serve as a valuable source of auxiliary information, thereby enhancing image understanding for applications such as image retrieval~\cite{johnson2015image,wang2020cross,yoon2021image}, image captioning~\cite{yang2019auto,li2019know,jia2023image}, image synthesis~\cite{johnson2018image,farshad2023scenegenie,wu2023scene}, and visual question answering~\cite{zhang2019empirical,qian2022scene,lei2023symbolic}.
The seminal work in this domain was conducted by Xu \textit{et al.}~\cite{xu2017scene}, which employs iterative message passing to generate visually grounded scene graphs. Subsequent to this pioneering work, several researchers  have adopted the message passing mechanism to better comprehend visual context~\cite{yang2018graph,gu2019scene,dai2017detecting,li2018factorizable,wang2019exploring}.

While traditional SGG approaches have shown promising results, they often suffer from the long-tailed distribution of relation predicates~\cite{dong2022stacked,li2022rethinking,he2022state,sun2023unbiased}.
Predicates in visual relations are often unevenly distributed, with head predicates (\textit{e.g.}, \ttt{on}, \ttt{have}) dominating the relation expressions~\cite{tang2020unbiased,li2021bipartite,yan2020pcpl,xu2022meta,han2022divide,jiang2023scene}. Such general relation expressions, however, offer limited utility for in-depth visual relation analysis~\cite{agarwal2020visual,guo2021general,goel2022not}. To address this challenge, He \textit{et al.}~\cite{he2021learning} introduces a knowledge transfer mechanism to leverage insights from head relations to enhance the representation of tail relations. 
Guo \textit{et al.}~\cite{guo2021general} refines biased predicate predictions based on the confusion matrix generated by training data.
Our work differs from conventional SGG in that we don’t assume that observations are perfect. We allow for corruptions in images, which are typical in real-world situations.

\looseness=-1
\textbf{Knowledge Based SGG.}
Recently, several approaches have been proposed to integrate external knowledge, referred to as \textit{commonsense}, to refine predicate and object prediction~\cite{bhagat2023sample,bhagat2023knowledge} and enhance the generalizability of the SGG model~\cite{gu2019scene,chen2019knowledge,zareian2020learning,ye2021linguistic,li2023label}. Specifically, GB-Net~\cite{zareian2020bridging} suggests that a scene graph can be perceived as an instantiation of a commonsense knowledge graph conditioned by the content of the image, and employs GGNN~\cite{li2016gated} to iteratively propagate messages between these two graphs for SGG task. Furthermore, EB-Net~\cite{chen2023more} advances this by enriching the knowledge graph for SGG with off-scene entities, thereby offering a more comprehensive and context-aware scene graph representation. In this work, we extend this by introducing superclass nodes and incorporating hierarchical edges into the knowledge graph, thereby facilitating hierarchical prediction for SGG models.
This is particularly advantageous when observations are corrupted, where features for specific classes are not easily detectable. In such cases, the hierarchical knowledge guides the model to first detect the superclass features. By adopting this approach, we can streamline the search space and facilitate more accurate predictions for finer classes.

\looseness=-1
\textbf{Corrupted Observation Perception.} In many computer vision tasks, it is a common assumption among researchers that the input image is invariably flawless and clear. However, this is often not the case in practical scenarios. To address this important issue, several benchmarks have been introduced to assess the robustness of the neural network models to real-world corruptions~\cite{mintun2021interaction,hendrycks2018benchmarking}. Within the context of corruption robustness, recent advancements can be broadly categorized into transfer learning~\cite{mirza2022norm,tang2023neuro,tang2023cross}, adversarial training~\cite{rusak2020simple,herrmann2022pyramid,kireev2022effectiveness}, data augmentation~\cite{zhang2018mixup,hendrycks2019augmix,yin2019fourier,zhang2022memo}, and large-scale pre-training~\cite{caron2021emerging,radford2021learning,fang2023corrupted}. 
Recently, LogicDef~\cite{yang2022logicdef} 
proposes a logic rules based defense method for adversarial patch attacks on images with multiple objects, utilizing logic rules learned from object relations to identify the attacked object. However, their approach assumes that the attack patch is on one single object, known to be under attack. Additionally, they assume that the relations between objects remain unaffected by the attack. In contrast, our work allows for corruption to occur at any location, potentially impacting an unknown number of objects and relations, which is more challenging as well as more realistic. 
To the best of our knowledge, ours is the first work to introduce corruptions into SGG and to propose the integration of hierarchical knowledge to ensure robust SGG in the presence of such corruptions.
\section{HiKER-SGG}
We introduce a novel framework HiKER-SGG, as illustrated in Figure~\ref{fig:overview}, to enable robust scene understanding for observations with potential corruptions.

\subsection{Problem Definition}
\looseness=-1
Given an image $\mathcal{I}$ in a dataset $\bm{\mathcal{I}}$, the SGG model aims to generate a directed scene graph $\mathcal{G}=\{\mathcal{N}, \mathcal{E}\}$, where each node $\mathcal{N}_i \in \mathcal{N}$ in the scene graph represents a localized object with bounding box $b_i$ and object class $\mathcal{C}_i^{\mathrm{E}}$, and each edge $\mathcal{E}_i \in \mathcal{E}$ denotes a predicate class $\mathcal{C}_i^{\mathrm{P}}$ between two objects. A well-constructed scene graph $\mathcal{G}$ contains a collection of visual relation triplets ($\langle$\ttt{subject}-\ttt{predicate}-\ttt{object}$\rangle$), which can be utilized to comprehensively describe the image $\mathcal{I}$. 

Our proposed HiKER-SGG follows a two-stage paradigm. We first generate a set of entity proposals with corresponding features using an off-the-shelf object detector (\textit{e.g.} Faster-RCNN~\cite{ren2015faster}) with a feature extraction network (\textit{e.g.} VGG~\cite{simonyan2015very} or ResNet~\cite{he2016deep}). The features extracted from the union box between two entities are used to represent their associated predicates.  Leveraging these features, we jointly make predictions for both the entity and predicate classes.

\subsection{Hierarchical Structure Discovery}
\label{sec:discover}
\looseness=-1
At the center of our work lies the hierarchical representation of domain knowledge. In this section, we introduce our automated approach to define hierarchies given GloVe~\cite{pennington2014glove} word embeddings and pattern similarity using MotifNet~\cite{zellers2018neural}.
A straightforward method is to manually set up these hierarchical relations. 
For instance, we can follow Zellers \textit{et al.}~\cite{zellers2018neural} to categorize 50 predicate classes into 3 superclasses, namely \ttt{geometric}, \ttt{possessive}, and \ttt{semantic}, respectively. Similarly, the 150 object classes can also be categorized into 12 superclasses, such as \ttt{artifact}, \ttt{animal}, \textit{etc}.


\looseness=-1
However, we recognize that there are various reasonable criteria for defining these hierarchies (\textit{e.g.}, by functions, sizes, materials). Setting up these hierarchies manually introduces subjectivity, which could hinder the capability of our approach on the unbiased SGG task. 
To address this issue, we adopt a hierarchical clustering~\cite{johnson1967hierarchical} algorithm, capable of revealing multi-level clusters based on a similarity metric, to discover the hierarchical structure for the entity and predicate classes. The similarity function used in hierarchical clustering is the weighted sum of the following two similarities:

\textit{(1) Semantic Similarity}. We use the GloVe~\cite{pennington2014glove} word embeddings $\mathbf{e}^{\mathrm{E}}$ and $\mathbf{e}^{\mathrm{P}}$ to calculate the cosine similarity between each pair of entities (E) and predicates (P):
\begin{equation}
\setlength\abovedisplayskip{4pt}
\setlength\belowdisplayskip{4pt}
    \mathcal{S}_{\mathrm{sem}}\left( \mathcal{C}_i^{\mathrm{E}/\mathrm{P}}, \mathcal{C}_j^{\mathrm{E}/\mathrm{P}}\right) = \frac{\mathbf{e}^{\mathrm{E}/\mathrm{P}}_i \cdot \mathbf{e}^{\mathrm{E}/\mathrm{P}}_j}{\left\| \mathbf{e}^{\mathrm{E}/\mathrm{P}}_i \right\|\left\| \mathbf{e}^{\mathrm{E}/\mathrm{P}}_j \right\|}.
\end{equation}

\looseness=-1
\textit{(2) Pattern Similarity}. We employ the MotifNet~\cite{zellers2018neural} baseline to generate confusion matrices $\mathcal{R}^{\mathrm{E}/\mathrm{P}}$ for both entities and predicates on the training dataset of Visual Genome~\cite{krishna2017visual}. Each matrix entry, $\mathcal{R}_{ij}$, indicates the likelihood (between 0 and 1) that the actual class is $i$ and the predicted class is $j$. Recognizing that similar classes often have similar patterns that might confuse our model, we compute the similarity based on the probability of the baseline method's misclassification between pairs of entities and predicates, written as
\begin{equation}
    \mathcal{S}_{\mathrm{pat}}\left( \mathcal{C}_i^{\mathrm{E}/\mathrm{P}}, \mathcal{C}_j^{\mathrm{E}/\mathrm{P}}\right) = \mathcal{R}_{ij}^{\mathrm{E}/\mathrm{P}} + \mathcal{R}_{ji}^{\mathrm{E}/\mathrm{P}}
\end{equation}

\looseness=-1
The hierarchies discovered through this method, which consider both semantic and pattern similarities, offer a more effective guidance for our hierarchical prediction approach, as discussed in Section \ref{sec:ablation}. More details about the clustering algorithm and hierarchy visualization can be found in Section A.1 of the Supplementary Materials.


\subsection{Hierarchical Knowledge Construction}
In the previous section, we discovered the hierarchies using those two metrics. This section details the representation of this hierarchical knowledge in our commonsense graph.

\textbf{Commonsense Knowledge Graph.}
Initially, we construct a commonsense knowledge graph that does not incorporate hierarchical knowledge.
Similar to GB-Net~\cite{zareian2020bridging}, we leverage
a {commonsense knowledge graph} which contains the possible relations between objects derived from extensive datasets like ConceptNet~\cite{speer2017conceptnet}, WordNet~\cite{miller1995wordnet}, \textit{etc}. Its edges serve as repositories of information regarding the general knowledge associated with objects, exemplified by connections such as \ttt{man}-\ttt{wears}-\ttt{shirt} and \ttt{cat}-\ttt{is}-\ttt{animal}. For simplicity, we define our commonsense graph as comprising a set of commonsense entity (CE) nodes $\mathcal{N}_\mathrm{CE}$ and commonsense predicate (CP) nodes $\mathcal{N}_\mathrm{CP}$ that are present in our SGG task.
%
The edges in the commonsense graph $\mathcal{E}_\mathrm{C}$ store the relations between each pair of nodes in both sets, which can be formally denoted as

\begin{small}
\vspace{-10pt}
\begin{equation}
\setlength\abovedisplayskip{3pt}
\setlength\belowdisplayskip{4pt}
\mathcal{E}_\mathrm{C} = \{\mathcal{E}^{\mathrm{CE}\rightarrow\mathrm{CP}}_{\texttt{relation}}\}\!\cup\!\{\mathcal{E}^{\mathrm{CP}\rightarrow\mathrm{CE}}_{\texttt{relation}}\}\!\cup\!\{\mathcal{E}^{\mathrm{CE}\rightarrow\mathrm{CE}}_{\texttt{relation}}\}\!\cup\!\{\mathcal{E}^{\mathrm{CP}\rightarrow\mathrm{CP}}_{\texttt{relation}}\}.
\end{equation}   
\end{small}

We initialize the CE and CP nodes features with a linear projection of their word embeddings~\cite{pennington2014glove} $\mathbf{e}^{\mathrm{E}}_i$ and $\mathbf{e}^{\mathrm{P}}_i$:
\begin{equation}
\mathbf{x}^{\mathrm{CE}}_i\!=\!\mathtt{LinearProj}(\mathbf{e}^{\mathrm{E}}_i)
,\,
\mathbf{x}^{\mathrm{CP}}_i\!=\!\mathtt{LinearProj}(\mathbf{e}^{\mathrm{P}}_i).
\end{equation}

\textbf{Hierarchical Commonsense Knowledge Graph.}
\looseness=-1
To  integrate hierarchical information discovered in Section \ref{sec:discover} into the prediction process, we introduce a set of specialized entity and predicate nodes across different levels within the commonsense knowledge graph, referred to as commonsense superclass entity (CXE) and commonsense superclass predicate (CXP) nodes\footnote{We use ``X" as the notation for ``superclass" to avoid ambiguity.}, as shown in Figure~\ref{fig:overview}. These nodes are denoted as $\mathcal{N}_\mathrm{CXE}$ and $\mathcal{N}_\mathrm{CXP}$, and correspond to a set of overarching categories for entities and predicates, respectively.

The initial representations of these superclass nodes are established by averaging the representations of $N_k$ subclass CE/CP nodes associated with each superclass, as follows:
\vspace{-2pt}
\begin{equation}
\hspace{-1mm}
\mathbf{x}^{\mathrm{CXE/CXP}}_{k}\!=\!\frac{\sum\nolimits_i \mathbf{x}_{i}^{\mathrm{CE/CP}}}{N_k}\!=\!\frac{\sum\nolimits_i \mathtt{LinearProj}(\mathbf{e}^{\mathrm{E/P}}_{i})}{N_k}.
\end{equation}
\vspace{-5pt}

We also establish binary connections $\mathcal{E}^{\mathrm{CXP}\rightarrow\mathrm{CP/CXP}}_{\mathtt{hierarchical}}$ and $\mathcal{E}^{\mathrm{CP/CXP}\rightarrow\mathrm{CXP}}_{\mathtt{hierarchical}}$ within the node sets $\mathcal{N}_\mathrm{CXP}$ and $\mathcal{N}_\mathrm{CP}$ to encode hierarchical information\footnote{In order to represent the multi-level hierarchy we discovered, two CXE/CXP nodes at different levels may also exhibit a hierarchical relation.}. Similar hierarchical edges are also established for the entity nodes. These edges also facilitate message passing, enabling the updating of superclass node representations, which are subsequently employed in computing superclass similarities. The final edges in the commonsense graph $\mathcal{E}_\mathrm{C}$ can be represented by
\vspace{-2pt}
\begin{align}
\mathcal{E}_\mathrm{C}\!= &\{\mathcal{E}^{\mathrm{CE}\rightarrow\mathrm{CP}}_{\mathtt{relation}}\}\!\cup\!\{\mathcal{E}^{\mathrm{CP}\rightarrow\mathrm{CE}}_{\mathtt{relation}}\}\!\cup\!\{\mathcal{E}^{\mathrm{CE}\rightarrow\mathrm{CE}}_{\mathtt{relation}}\} \cup \nonumber\\&\{\mathcal{E}^{\mathrm{CP}\rightarrow\mathrm{CP}}_{\mathtt{relation}}\}\!\cup\!\{\mathcal{E}^{\mathrm{CXE}\rightarrow\mathrm{CE/CXE}}_{\mathtt{hierarchical}}\}\!\cup\!\{\mathcal{E}^{\mathrm{CE/CXE}\rightarrow\mathrm{CXE}}_{\mathtt{hierarchical}}\}\cup\nonumber\\&\{\mathcal{E}^{\mathrm{CXP}\rightarrow\mathrm{CP/CXP}}_{\mathtt{hierarchical}}\}\!\cup\!\{\mathcal{E}^{\mathrm{CP/CXP}\rightarrow\mathrm{CXP}}_{\mathtt{hierarchical}}\}.
\end{align}
\vspace{-20pt}

\subsection{Scene Graph Initialization}
\looseness=-1
So far, we developed a hierarchical commonsense knowledge graph sourced from knowledge databases. Our next step is to construct a scene graph from the given input image.

A scene graph is different from a commonsense graph in that: 
(1) each scene entity (SE) node $\mathcal{N}_\mathrm{SE}$ is associated with a bounding box, \textit{i.e.} $\mathcal{N}_\mathrm{SE} \subseteq [0,1]^4 \times \mathcal{N}_\mathrm{CE}$; 
(2) each scene predicate (SP) node $\mathcal{N}_\mathrm{SP}$ is associated with a pair of SE nodes, \textit{i.e.} $\mathcal{N}_\mathrm{SP} \subseteq \mathcal{N}_\mathrm{SE} \times \mathcal{N}_\mathrm{SE} \times \mathcal{N}_\mathrm{CP}$. The directed edges $\mathcal{E}_\mathrm{S}$ in the scene graph can be similarly defined as

\begin{small}
\vspace{-12pt}
\begin{equation}
    \mathcal{E}_\mathrm{S} = \{\mathcal{E}^{\mathrm{SE}\rightarrow\mathrm{SP}}_\mathtt{subjectOf}\} \cup \{\mathcal{E}^{\mathrm{SE}\rightarrow\mathrm{SP}}_\mathtt{objectOf}\} \cup \{\mathcal{E}^{\mathrm{SP}\rightarrow\mathrm{SE}}_\mathtt{hasSubject}\} \cup \{\mathcal{E}^{\mathrm{SP}\rightarrow\mathrm{SE}}_\mathtt{hasObject}\}.
\end{equation}
\end{small}

In our SGG settings, the true classes for the SE/SP nodes might not be provided, which requires us to predict them. Therefore, we modify the scene graph entity nodes needed to be classified as $\mathcal{N}_\mathrm{SE}^{\mathsf{unk}} \subseteq  \, [0,1]^4$, and scene graph predicate nodes needed to be classified as $\mathcal{N}_\mathrm{SP}^{\mathsf{unk}} \subseteq \, \mathcal{N}_\mathrm{SE} \times \mathcal{N}_\mathrm{SE}$, where $\mathcal{N}_\mathrm{SE/SP}^{\mathsf{unk}}$ means the classes of the SE/SP nodes are unknown.

To initialize the scene graph for each sample, we first utilize the object detector to find potential objects. We then create a SE node for each object and a SP node for each pair of objects.  The SE node is initialized by RoI-aligned~\cite{ren2015faster} feature vector $\mathbf{v}^{\mathrm{E}}_{i}$, and the SP node is initialized by RoI feature $\mathbf{v}^{\mathrm{P}}_{i}$ of the union bounding box:
\begin{equation}
\setlength\abovedisplayskip{3pt}
\setlength\belowdisplayskip{4pt}
\label{fcn1}
\mathbf{x}^{\mathrm{SE}}_i = \mathtt{FCNet}(\mathbf{v}^{\mathrm{E}}_{i})
\;,\quad
\mathbf{x}^{\mathrm{SP}}_i = \mathtt{FCNet}(\mathbf{v}^{\mathrm{P}}_{i}),
\end{equation}
where \ttt{FCNet} denotes a fully connected network. 
It should be noted that the weights for these two fully connected networks are distinct and not shared. 

\subsection{Bridging Hierarchical Knowledge and SGG}

To bridge the knowledge graph and the scene graph, we create \textit{bridge edges} $\mathcal{E}_\mathrm{B}$ to facilitate the mutual information flow during training. Specifically, these bi-directional bridge edges link an entity or predicate from the scene graph to its corresponding labels in the commonsense graph\footnote{Given the symmetric nature of the relation,  the bridge edges are implemented as bi-directional directed edges with shared weights.}. The bridge edges $\mathcal{E}_\mathrm{B}$ can be defined as

\begin{small}
\vspace{-10pt}
\begin{equation}
\setlength\abovedisplayskip{3pt}
\setlength\belowdisplayskip{4pt}
    \mathcal{E}_\mathrm{B}\!=\!\{\mathcal{E}^{\mathrm{SE}\rightarrow\mathrm{CE}}_\mathtt{classTo}\}\cup\{\mathcal{E}^{\mathrm{SP}\rightarrow\mathrm{CP}}_\mathtt{classTo}\}\cup\{\mathcal{E}^{\mathrm{CE}\rightarrow\mathrm{SE}}_\mathtt{hasInst}\}\cup\{\mathcal{E}^{\mathrm{CP}\rightarrow\mathrm{SP}}_\mathtt{hasInst}\}.
\end{equation}
\end{small}

Initially, we link each SE node to multiple CE nodes and assign weights based on the labels predicted by Faster R-CNN. The edges between SP and CP nodes start as an empty set and will be updated during message propagation.
Enforcing the information flow between the knowledge graph and the scene graph, we adopt a variant of GGNN~\cite{li2016gated} to update node representations and propagate messages among nodes using a Gated Recurrent Unit (GRU)~\cite{cho2014learning} updating rule: 
\begin{equation}
\setlength\abovedisplayskip{3pt}
\setlength\belowdisplayskip{4pt}
    \mathbf{x}_i^{\phi} \leftarrow \mathtt{GRU\,Update}(\mathbf{x}_i^{\phi}),
\end{equation}
where $\leftarrow$ denotes updating the node representation, with the superscript $\phi \in \{\mathrm{SE, SP, CE, CP, CXE, CXP}\}$.

\looseness=-1
After each iteration of message propagation, we compute the similarities of each SE/SP node to all CE/CP nodes by 
\begin{equation} 
\setlength\abovedisplayskip{3pt}
\setlength\belowdisplayskip{4pt}
\label{fcn2}
\mathrm{sim}\left(\mathbf{x}_i^\phi,\mathbf{x}_j^\phi \right)= \left(\mathtt{FCNet}\left(\mathbf{x}_i^\phi\right)\right) ^\top \left(\mathtt{FCNet}\left(\mathbf{x}_j^\phi\right)\right).
\end{equation}
The pairwise similarities, which quantify the connections between scene nodes and commonsense nodes, are used to update the weights of the bridge edges after each iteration.
Explicitly, the weights of the bridge edges $\mathcal{E}_\mathrm{B}$ are updated by
\begin{align}
\mathbf{w}_{ij}^{\mathrm{SE}\leftrightarrow \mathrm{CE}} &\leftarrow \frac{\exp\!\left( \mathrm{sim}\!\left(\mathbf{x}_i^\mathrm{SE},\mathbf{x}_j^\mathrm{CE} \right) \right)}{\sum\nolimits_{j'}{\exp\!\left( \mathrm{sim}\!\left(\mathbf{x}_i^\mathrm{SE},\mathbf{x}_{j'}^\mathrm{CE} \right) \right)}}, \\
\mathbf{w}_{ij}^{\mathrm{SP}\leftrightarrow \mathrm{CP}} &\leftarrow \frac{\exp\!\left( \mathrm{sim}\!\left(\mathbf{x}_i^\mathrm{SP},\mathbf{x}_j^\mathrm{CP} \right) \right)}{\sum\nolimits_{j'}{\exp\!\left( \mathrm{sim}\!\left(\mathbf{x}_i^\mathrm{SP},\mathbf{x}_{j'}^\mathrm{CP} \right) \right)}},
\end{align}
where $\mathbf{w}_{ij}^{\mathrm{SE}\leftrightarrow \mathrm{CE}}$ and $\mathbf{w}_{ij}^{\mathrm{SP}\leftrightarrow \mathrm{CP}}$ represent the shared weights of bi-directional bridge edges connecting a specific pair of SE/SP and CE/CP nodes, respectively.
After $t$ steps of message propagation, we can leverage the node representations from both graphs to infer the unknown class of SE/SP nodes.

\subsection{Hierarchical Inference}
Using the updated node representations in both graphs, we propose to determine the class of each unknown SE/SP node by a hierarchical inference process.
Here, we present the inference process for predicate classification only. The same paradigm is also applied to entity nodes. 

\looseness=-1
Specifically, We enforce our model to infer the predicate class sequentially from higher to lower levels. For simplicity, we introduce our approach using a 3-level hierarchy; however, this hierarchical inference scheme is scalable to accommodate a more complex hierarchy. In the 3-level case, the CXP nodes can be split into two groups: higher-level nodes denoted by $\mathcal{N}_\mathrm{CXP}^{(1)}$ and lower-level nodes denoted by $\mathcal{N}_\mathrm{CXP}^{(2)}$. The hierarchical path from the top superclass node to the final subclass node can be expressed as $\mathcal{N}_\mathrm{CXP}^{(1)}\!\rightarrow\!\mathcal{N}_\mathrm{CXP}^{(2)}\!\rightarrow\!\mathcal{N}_\mathrm{CP}$, which corresponds to the classification sequence from higher to lower predicate class: $\mathcal{C}^{\mathrm{XP1}}\!\rightarrow\!\mathcal{C}^{\mathrm{XP2}}\!\rightarrow\! \mathcal{C}^{\mathrm{P}}$.

Specifically, we first compute the similarities between the node representations of each SP node and the higher-level CXP nodes within the hierarchical knowledge graph to determine the level-1 superclass probabilities, written as
\begin{equation}
\setlength\abovedisplayskip{5pt}
\setlength\belowdisplayskip{5pt}
\label{eq:superP}
\mathbb{P}\!\left(\mathcal{C}^{\mathrm{XP1}}\, | \mathcal{N}_\mathrm{SP}^{\mathsf{unk}}\right) = \mathtt{Softmax}\left( \mathrm{sim}\!\left( \mathbf{x}_i^\mathrm{SP}, \mathbf{x}^{\mathrm{CXP1}}_{k_1}\right)\right).
\end{equation}
Here, $k_1$ denotes the level-1 superclass indices, $\mathbf{x}^{\mathrm{CXP1}}_{k_1}$ denotes the node representation for $\mathcal{N}_\mathrm{CXP}^{(1)}$, and $\mathrm{sim}(\cdot, \cdot)$ is defined according to Equation (\ref{fcn2}). 

Once we have classified the level-1 superclass for each unknown predicate node in the scene graph, we then examine the conditional probabilities $\mathbb{P}\!\left(\mathcal{C}^{\mathrm{XP2}} | \mathcal{N}_\mathrm{SP}^{\mathsf{unk}}, \mathcal{C}^{\mathrm{XP1}}\right)$, \textit{i.e.}, the probabilities of level-2 superclass predicates given the level-1 superclass. The probabilities can be computed as follows:
\begin{equation}
\setlength\abovedisplayskip{5pt}
\setlength\belowdisplayskip{5pt}
\hspace{-2.5mm}
\mathbb{P}\!\left(\mathcal{C}^{\mathrm{XP2}} | \mathcal{N}_\mathrm{SP}^{\mathsf{unk}}, \mathcal{C}^{\mathrm{XP1}}\right)\!=\!\mathtt{Softmax}\left(\mathrm{sim}\!\left( \mathbf{x}_i^\mathrm{SP}\!, \mathbf{x}^{\mathrm{CXP2}}_{k_2}\right)\right)\!,
\end{equation}
where $k_2$ denotes the level-2 superclass predicate indices in a given level-1 superclass. 
Ultimately, the conditional probabilities of final subclass predicates can be written as
\begin{equation}
\setlength\abovedisplayskip{5pt}
\setlength\belowdisplayskip{5pt}
\mathbb{P}\!\left( \mathcal{C}^{\mathrm{P}} | \mathcal{N}_\mathrm{SP}^{\mathsf{unk}},  \mathcal{C}^{\mathrm{XP2}}\right) = \mathtt{Softmax}\left( \mathrm{sim}\!\left( \mathbf{x}_i^\mathrm{SP}, \mathbf{x}^{\mathrm{CP}}_{j}\right)\right).
\end{equation}

In general, given an unknown predicate node, the predicted probability of each predicate category can be computed by multiplying the three probabilities derived above:
\begin{flalign}
\label{eq:pred}
&\mathbb{P}\!\left(\mathcal{C}^{\mathrm{P}} | \mathcal{N}_\mathrm{SP}^{\mathsf{unk}}\right)\!= \mathbb{P} \!\left( \mathcal{C} ^{\mathrm{XP}2}|\mathcal{N} _{\mathrm{SP}}^{\mathsf{unk}} \right) \cdot \mathbb{P} \!\left( \mathcal{C} ^{\mathrm{P}}|\mathcal{N}_{\mathrm{SP}}^{\mathsf{unk}},\mathcal{C} ^{\mathrm{XP}2} \right)\\
&=\!
\mathbb{P}\!\left(\mathcal{C}^{\mathrm{XP1}}| \mathcal{N}_\mathrm{SP}^{\mathsf{unk}}\right)
\!\cdot \mathbb{P}\!\left(\mathcal{C}^{\mathrm{XP2}} | \mathcal{N}_\mathrm{SP}^{\mathsf{unk}}\!, \mathcal{C}^{\mathrm{XP1}}\right)\!\cdot \mathbb{P}\!\left( \mathcal{C}^{\mathrm{P}} | \mathcal{N}_\mathrm{SP}^{\mathsf{unk}}\!,  \mathcal{C}^{\mathrm{XP2}}\right)\!. \nonumber
\end{flalign}

\subsection{Adaptive Refinement}
Due to the inherent bias in the Visual Genome~\cite{krishna2017visual} dataset, most existing SGG models tend to favor commonly occurring predicate classes. In this work, we integrate an adaptive refinement mechanism into our model to mitigate biases in predicate classes. This enhancement aims to predict more specific and informative predicates (\textit{e.g.},  \ttt{standing} \ttt{on}, \ttt{sitting} \ttt{on}), as opposed to general ones (\textit{e.g.}, \ttt{on}). Essentially, our goal is to find transitioning probabilities $\mathbb{P}( \mathcal{C}_s^{\mathrm{P}} | \mathcal{C}_g^{\mathrm{P}})$  that can convert a general prediction into a more specific prediction for predicate classes.

Unlike previous method like G2S~\cite{guo2021general} which incorporates fixed transitioning probabilities to debias the predictions, our adaptive refinement dynamically updates the transition probabilities during the training process.
Specifically, we adopt the predicate confusion matrix generated by the MotifNet~\cite{zellers2018neural} baseline as initialization for $\mathcal{R}$.
We then create a transitioning probability matrix by row-normalizing the diagonal-augmented confusion matrix:
\begin{equation}
\setlength\abovedisplayskip{5pt}
\setlength\belowdisplayskip{5pt}
\label{eq:transition}
    \mathcal{T} = \mathtt{RowNormalize}\left(\mathcal{R}+I\right),
\end{equation}
where $I$ represents an identity matrix of the same size as the confusion matrix $\mathcal{R}$. The transitioning probability $\mathbb{P}( \mathcal{C}_s^{\mathrm{P}} | \mathcal{C}_g^{\mathrm{P}})$ can be subsequently represented by a particular entry $\mathcal{T}_{ij}$, which aligns with the respective classes $\mathcal{C}_s^{\mathrm{P}}$ and $\mathcal{C}_g^{\mathrm{P}}$.

Combining this refinement with our hierarchical prediction approach,  we can rewrite Equation (\ref{eq:pred}) as:
\vspace{-3pt}
\begin{align}
\label{eq:finalP}
\mathbb{P}\!\left(\mathcal{C}^{\mathrm{P}} | \mathcal{N}_\mathrm{SP}^{\mathsf{unk}}\right) =& 
\mathbb{P}\!\left(\mathcal{C}^{\mathrm{XP1}}\, | \mathcal{N}_\mathrm{SP}^{\mathsf{unk}}\right)
\cdot \mathbb{P}\!\left(\mathcal{C}^{\mathrm{XP2}} | \mathcal{N}_\mathrm{SP}^{\mathsf{unk}}, \mathcal{C}^{\mathrm{XP1}}\right)\nonumber\\
&\cdot \mathbb{P}\!\left( \mathcal{C}^{\mathrm{P}} | \mathcal{N}_\mathrm{SP}^{\mathsf{unk}},  \mathcal{C}^{\mathrm{XP2}}\right) \cdot \mathbb{P}\!\left( \mathcal{C}_s^{\mathrm{P}} | \mathcal{C}_g^{\mathrm{P}}\right).
\end{align} 
\vspace{-16pt}

During the training stage, we aim to uncover deeper correlations between predicate classes, facilitating a more fine-grained prediction.
Therefore, we propose to re-evaluate our SGG model on the training dataset  after each training epoch to obtain a new $\mathcal{T}^{m}$ following Equation (\ref{eq:transition}). We then blend this matrix with the one from the previous epoch using a weighted linear combination:
\begin{equation}
\label{eq:update}
\setlength\abovedisplayskip{5pt}
\setlength\belowdisplayskip{5pt}
    \mathcal{T}^{m} \leftarrow \alpha \mathcal{T}^{m} + (1-\alpha) \mathcal{T}^{m-1},
\end{equation}
where $m$ represents the current epoch index, and $\alpha$ is a hyperparameter to control the update rate. This updated matrix will be used for predicate classification in the next training epoch.
Additional discussions on adaptive refinement are provided in Section A.3 of the Supplementary Materials.

During the training stage, we update our parameters using the following loss terms to supervise both the superclass and subclass predictions defined in Equations (\ref{eq:superP}) and (\ref{eq:finalP}):
\vspace{-2pt}
\begin{align}
\mathcal{L}_{\mathrm{XP1}} &= \mathtt{NLL\,Loss}\!\left(\mathbb{P}\!\left(\mathcal{C}^{\mathrm{XP1}} | \mathcal{N}_\mathrm{SP}^{\mathsf{unk}}\right), \mathtt{OneHot}\!\left(\mathcal{C}^{\mathrm{XP1}}_{\mathrm{GT}}\right)\right),\nonumber\\
\mathcal{L}_{\mathrm{XP2}} &= \mathtt{NLL\,Loss}\!\left(\mathbb{P}\!\left(\mathcal{C}^{\mathrm{XP2}} | \mathcal{N}_\mathrm{SP}^{\mathsf{unk}}\right), \mathtt{OneHot}\!\left(\mathcal{C}^{\mathrm{XP2}}_{\mathrm{GT}}\right)\right),\nonumber\\
\mathcal{L}_{\mathrm{P}} &= \mathtt{NLL\,Loss}\!\left(\mathbb{P}\!\left(\mathcal{C}^{\mathrm{P}}| \mathcal{N}_\mathrm{SP}^{\mathsf{unk}}\right), \mathtt{OneHot}\!\left(\mathcal{C}^{\mathrm{P}}_{\mathrm{GT}}\right)\right),\\[-19pt]\nonumber
\end{align}
where $\mathcal{C}^{\mathrm{XP1/XP2}}_{\mathrm{GT}}$ and $\mathcal{C}^{\mathrm{P}}_{\mathrm{GT}}$ represent the ground-truth labels for the superclass and subclass predicates, respectively.

\section{Experiments}
In this section, we conduct extensive experiments on the large-scale Visual Genome (VG)~\cite{krishna2017visual} dataset and our corrupted Visual Genome (VG-C) benchmark. The results indicate that HiKER-SGG excels beyond state-of-the-art models with superior performance on both clean and corrupted images. It is noteworthy that our method is corruption-agnostic, as it is trained solely on clean images and directly tested on corrupted ones without additional training.

\begin{table*}[t]
\renewcommand\arraystretch{0.95}
\centering
\caption{\textbf{Performance comparison with the state-of-the-art SGG methods on the Visual Genome \cite{krishna2017visual} dataset}. The best results for each metric are in \textbf{bold}, while the second-best results are \underline{underlined}. ``-" denotes unavailable results due to incompatible experimental settings.}
\vspace{-8pt}
\label{tab:performance}
\resizebox{\linewidth}{!}{
\begin{tabular}{lccccccc}
\toprule
\multirow{2}{*}[-0.5ex]{Method}   &\multirow{2}{*}[-0.5ex]{Venue}   & \multicolumn{3}{c}{PredCls}              & \multicolumn{3}{c}{SGCls}                \\
\cmidrule(lr){3-5}\cmidrule(lr){6-8}
                       &     & mR@20: UC/C & mR@50: UC/C & mR@100: UC/C & mR@20: UC/C & mR@50: UC/C & mR@100: UC/C \\
\midrule
IMP+ \cite{xu2017scene} &\textit{CVPR'17}                         & - / -       & 20.3 / 9.8    & 28.9 / 10.5    &     - / -        &      12.1 / 9.8       &       16.9 / 10.5       \\
Neural Motifs \cite{zellers2018neural}   &\textit{CVPR'18}         & - / 10.8     & 24.8 / 14.0  & 37.3 / 15.3    &      - / 6.3       &    13.5 / 7.7        &        19.6 / 8.2     \\
VCTree  \cite{tang2019learning}       &\textit{CVPR'19}            & - / 14.0     & - / 17.9     & - / 19.4      &      - / 8.2      &      - / 10.1       &        - / 10.8      \\
PCPL    \cite{yan2020pcpl}            &\textit{ACMMM'20}            & - / -       & 50.6 / 35.2   & 62.6 / 37.8   &     - / -       &       \underline{26.8} / 18.6      &       \underline{32.8} / 19.6       \\
Transformer + CogTree \cite{yu2020cogtree} &\textit{IJCAI'21}        & - / 22.9     & - / 28.4     & - / 31.0      &      - / 13.0      &     - / 15.7        &         - / 16.7     \\
VCTree + EBM \cite{suhail2021energy}  &\textit{CVPR'21}     & - / 14.2     & - / 18.0     & - / 28.8      &       - / 8.2     &     - / 10.2        &         - / 11.0     \\
G2S: Transformer \cite{guo2021general}   &\textit{ICCV'21}         & - / 26.7     & - / 31.9     & - / 34.2      &      - / 15.7      &      - / 18.5       &         - / 19.4     \\
MotifNet + DLFE  \cite{chiou2021recovering}  &\textit{ACMMM'21}     & - / 22.1     & - / 26.9     & - / 28.8      &       - / 12.8     &     - / 15.2        &         - / 15.9     \\
MotifNet + RTPB \cite{chen2022resistance}   &\textit{AAAI'22}                     & - / 28.8     & - / 35.3     & - / 37.7      &      - / 16.3      &     - / 19.4        &         - / 20.6     \\
MotifNet + PPDL  \cite{li2022ppdl}   &\textit{CVPR'22}    & - / 27.9     & - / 32.2     & - / 33.3      &       - / 15.8     &     - / 17.5        &         - / 18.2     \\
MotifNet + NICE  \cite{li2022devil}   &\textit{CVPR'22}    & - / 23.7     & - / 29.8     & - / 32.2      &       - / 13.6     &     - / 16.7       &         - / 17.9     \\
MotifNet + NARE  \cite{goel2022not}   &\textit{CVPR'22}    & - / 21.3     & - / 27.1     & - / 29.7      &       - / 11.3     &     - / 14.3       &         - / 15.7     \\
Transformer + HML \cite{deng2022hierarchical}   &\textit{ECCV'22}                     & - / 27.4     & - / 33.3     & - / 35.9      &      - / 15.7      &     - / 19.1        &         - / 20.4     \\
SQUAT \cite{jung2023devil}                &\textit{CVPR'23}        & - / 25.6     & - / 30.9     & - / 33.4      &      - / 14.4      &     - / 17.5        &         - / 18.8     \\
PE-Net \cite{zheng2023prototype}                &\textit{CVPR'23}        & - / 25.8     & - / 31.4     & - / 33.5      &      - / 15.2      &     - / 18.2        &         - / 19.3     \\
PE-Net + SIL \cite{wang2023improving}                &\textit{ACMMM'23}        & - / 26.9     & - / 33.1     & - / 35.3      &      - / \underline{16.7}      &     - / \underline{19.9}        &         - / \underline{20.7}     \\
\midrule
GB-Net \cite{zareian2020bridging}     &\textit{ECCV'20}            & 23.8 / 15.3  & 41.1 / 19.3   & 55.4 / 20.9   &     13.1 / 7.9     &    21.4 / 9.6      &         29.1 / 10.2     \\
EB-Net + EOA \cite{chen2023more}     &\textit{WACV'23}             & \underline{39.8} / \underline{30.8}   & \underline{54.9} / \underline{36.7}     & \underline{66.3} / \underline{39.2}  &    \underline{19.6} / 14.9    &      26.7 / 17.3       &       32.5 / 18.3       \\
\rowcolor{gray!20}
\textbf{HiKER-SGG (Ours)}              &-                       &  \textbf{42.1} / \textbf{33.4}  & \textbf{57.9} / \textbf{39.3}  &  \textbf{69.2} / \textbf{41.2} &    \textbf{22.6} / \textbf{18.2}         &   \textbf{30.0} / \textbf{20.3}        &       \textbf{36.7} / \textbf{21.4}       \\
\bottomrule
\end{tabular}
}
\end{table*}

\begin{table*}[htbp]
\renewcommand\arraystretch{1}
\setlength{\abovecaptionskip}{0cm}
\setlength{\belowcaptionskip}{-0.2cm}
\begin{center}
\caption{\textbf{Performance comparison with the state-of-the-art SGG methods for the PredCls task on the corrupted Visual Genome \cite{krishna2017visual} dataset}.  We report the accuracy in percentage for the mR@20: UC/C, mR@50: UC/C, mR@100: UC/C metrics, structured in six rows. The best results for each metric are in \textbf{bold}. The last column reports the average mean recall across all 20 types of corruption, and the percentage decrease in \textcolor{MidnightBlue}{blue} when compared to the mean recall on clean images. $^\dag$We evaluate these methods using the codes provided by the authors.}
\label{tab:performance2}
\resizebox{\linewidth}{!}{
\begin{tabular}{c|l|cccccccccccccccccccc|cc}
\toprule
& Method & gaus & shot & imp & dfcs & gls & mtn & zm & snw & frst & fg & brt & cnt & els & px & jpg & sun & wtd & smk & rain & dust & Average mR \\			
\midrule
\multirow{6}{*}[-0.75ex]{\centering \rotatebox{90}{mR@20: C/UC}}&GB-Net$^\dag$~\cite{zareian2020bridging} & 15.2 & 16.0 & 15.2 & 16.9 & 14.9 & 16.5 & 16.6 & 17.9 & 18.9 & 21.4 & 21.6 & 14.7 & 16.8 & 16.6 & 18.2 & 16.7 & 17.8 & 16.0 & 20.1 & 18.5 & 17.3 (\textcolor{MidnightBlue}{-27.3\%})\\
&EB-Net$^\dag$~\cite{chen2023more}        & 28.0 & 29.8 & 27.4 & 31.2 & 26.5 & 30.3  & 30.5  & 32.1 & 33.2 & 35.8 & 36.3 & 27.3 & 30.3 & 27.0 & 30.6 & 30.6 & 30.7 & 33.7 & 35.6& 30.1 & 30.9 (\textcolor{MidnightBlue}{-22.4\%})\\ 
&\cc\textbf{HiKER-SGG} & \cc\textbf{31.1} & \cc\textbf{33.3} & \cc\textbf{31.5} & \cc\textbf{35.4} & \cc\textbf{28.5}  & \cc\textbf{35.0}  & \cc\textbf{34.1}  & \cc\textbf{36.5} & \cc\textbf{37.7} & \cc\textbf{39.8} & \cc\textbf{40.8} & \cc\textbf{30.5} & \cc\textbf{33.7} & \cc\textbf{31.3} & \cc\textbf{34.2} & \cc\textbf{33.5} & \cc\textbf{34.9} & \cc\textbf{37.1} & \cc\textbf{39.8} & \cc\textbf{32.6} & \cc\textbf{34.6} (\textcolor{MidnightBlue}{\cc\textbf{-17.8\%}})\\
\cmidrule{2-23}
&GB-Net$^\dag$~\cite{zareian2020bridging} & 10.3 & 10.6 & 10.4 & 11.6 & 10.4 & 10.9 & 10.7 & 11.9 & 12.3 & 13.7 & 13.8 & 10.0 & 11.1 & 10.8 & 11.7 & 11.1 & 11.2 & 10.5 & 13.0 & 12.1 & 11.4 (\textcolor{MidnightBlue}{-25.5\%})\\
&EB-Net$^\dag$~\cite{chen2023more}        & 21.7 & 22.8 & 20.4 & 24.9 & 19.6 & 23.2 & 23.8 & 23.2 & 24.6 & 27.5 & 28.0 & 20.1 & 23.1 & 21.1 & 23.6 & 24.0 & 23.4 & 25.6 & 27.3 & 22.9 & 23.5 (\textcolor{MidnightBlue}{-23.7\%})\\
&\cc\textbf{HiKER-SGG} & \cc\textbf{24.8} & \cc\textbf{25.8} & \cc\textbf{24.8} & \cc\textbf{27.5} & \cc\textbf{22.4} & \cc\textbf{27.4} & \cc\textbf{26.4} & \cc\textbf{27.8} & \cc\textbf{28.7} & \cc\textbf{31.1} & \cc\textbf{31.5} & \cc\textbf{23.3} & \cc\textbf{26.0} & \cc\textbf{24.3} & \cc\textbf{26.5} & \cc\textbf{26.3} & \cc\textbf{26.8} & \cc\textbf{28.5} & \cc\textbf{30.9} & \cc\textbf{24.9} & \cc\textbf{26.8} (\textcolor{MidnightBlue}{\cc\textbf{-19.8\%}})\\
\midrule
\multirow{6}{*}[-0.75ex]{\centering \rotatebox{90}{mR@50: C/UC}}&GB-Net$^\dag$~\cite{zareian2020bridging} & 27.5 & 28.7 & 27.6 & 30.8 & 26.4 & 29.8 & 29.9 & 31.9 & 33.8 & 37.2 & 37.6 & 26.3 & 29.9 & 30.0 & 33.0 & 29.5 & 32.3 & 28.7 & 35.8 & 32.8 & 31.0 (\textcolor{MidnightBlue}{-24.6\%})\\
&EB-Net$^\dag$~\cite{chen2023more}        & 42.1 & 43.7 & 41.5 & 44.9 & 40.2 & 45.6 & 44.2 & 46.9 & 47.7 & 50.4 & 51.2 & 41.2 & 44.1 & 41.4 & 45.1 & 45.4 & 45.5 & 48.4 & 49.7 & 44.6 & 45.2 
 (\textcolor{MidnightBlue}{-17.7\%})\\ 
&\cc\textbf{HiKER-SGG} & \cc\textbf{46.7} & \cc\textbf{48.4} & \cc\textbf{46.9} & \cc\textbf{50.2} & \cc\textbf{43.2} & \cc\textbf{49.6} & \cc\textbf{48.3} & \cc\textbf{51.3} & \cc\textbf{52.5} & \cc\textbf{55.1} & \cc\textbf{55.9} & \cc\textbf{45.0} & \cc\textbf{48.1} & \cc\textbf{46.0} & \cc\textbf{49.9} & \cc\textbf{48.6} & \cc\textbf{50.0} & \cc\textbf{52.4} & \cc\textbf{54.8} & \cc\textbf{47.0} & \cc\textbf{49.5} 
 (\textcolor{MidnightBlue}{\cc\textbf{-14.5\%}})\\
\cmidrule{2-23}
&GB-Net$^\dag$~\cite{zareian2020bridging} & 13.3 & 13.6 & 13.3 & 15.1 & 13.6 & 14.1 & 14.0 & 15.4 & 15.6 & 17.4 & 17.5 & 13.0 & 14.5 & 14.4 & 15.2 & 14.5 & 14.6 & 13.6 & 16.6 & 15.4 & 14.7 
 (\textcolor{MidnightBlue}{-24.2\%})\\
&EB-Net$^\dag$~\cite{chen2023more}        & 24.8 & 27.6 & 25.6 & 28.3 & 25.9 & 28.9 & 29.4 & 29.3 & 30.5 & 32.0 & 32.8 & 26.1 & 28.6 & 26.3 & 27.9 & 29.2 & 28.6 & 30.8 & 31.8 & 27.2 & 28.6 
(\textcolor{MidnightBlue}{-22.1\%})\\ 
&\cc\textbf{HiKER-SGG} & \cc\textbf{30.1} & \cc\textbf{31.7} & \cc\textbf{30.4} & \cc\textbf{33.2} & \cc\textbf{28.3} & \cc\textbf{33.3} & \cc\textbf{32.1} & \cc\textbf{34.1} & \cc\textbf{34.4} & \cc\textbf{37.3} & \cc\textbf{37.4} & \cc\textbf{28.8} & \cc\textbf{31.7} & \cc\textbf{30.1} & \cc\textbf{32.9} & \cc\textbf{32.5} & \cc\textbf{32.2} & \cc\textbf{34.5} & \cc\textbf{36.7} & \cc\textbf{30.4} & \cc\textbf{32.6} 
 (\textcolor{MidnightBlue}{\cc\textbf{-17.0\%}})\\
\midrule
\multirow{6}{*}[-0.3ex]{\centering \rotatebox{90}{mR@100: C/UC}}&GB-Net$^\dag$~\cite{zareian2020bridging} & 40.1 & 41.9 & 40.1 & 43.8 & 37.8 & 42.9 & 42.7 & 45.1 & 47.1 & 50.8 & 51.7 & 37.8 & 42.8 & 42.9 & 46.6 & 42.5 & 46.1 & 41.2 & 49.6 & 45.9 & 44.0 
 (\textcolor{MidnightBlue}{-20.6\%})\\
&EB-Net$^\dag$~\cite{chen2023more}        & 54.7 & 56.0 & 52.9 & 56.8 & 52.4 & 55.6 & 55.3 & 58.4 & 59.9 & 61.6 & 61.1 & 53.3 & 55.0 & 54.3 & 57.7 & 56.4 & 57.6 & 59.0 & 60.7 & 54.8 & 56.7 
 (\textcolor{MidnightBlue}{-14.5\%})\\
&\cc\textbf{HiKER-SGG} & \cc\textbf{59.3} & \cc\textbf{60.3} & \cc\textbf{58.6} & \cc\textbf{62.3} & \cc\textbf{55.6} & \cc\textbf{61.9} & \cc\textbf{59.8} & \cc\textbf{63.4} & \cc\textbf{64.0} & \cc\textbf{66.9} & \cc\textbf{67.4} & \cc\textbf{56.4} & \cc\textbf{60.1} & \cc\textbf{58.4} & \cc\textbf{62.3} & \cc\textbf{59.8} & \cc\textbf{62.1} & \cc\textbf{63.7} & \cc\textbf{66.3} & \cc\textbf{58.9} & \cc\textbf{61.4} 
 (\textcolor{MidnightBlue}{\cc\textbf{-11.3\%}})\\
\cmidrule{2-23}
&GB-Net$^\dag$~\cite{zareian2020bridging} & 14.8 & 15.1 & 14.6 & 16.6 & 15.1 & 15.6 & 15.6 & 16.9 & 17.1 & 19.1 & 19.0 & 14.4 & 16.0 & 16.0 & 16.8 & 16.1 & 16.1 & 15.0 & 18.1 & 17.0 & 16.3 
(\textcolor{MidnightBlue}{-22.0\%})\\
&EB-Net$^\dag$~\cite{chen2023more}        & 28.7 & 30.1 & 27.8 & 31.9 & 27.1 & 31.1 & 30.5 & 32.8 & 32.4 & 36.1 & 35.7 & 28.2 & 30.9 & 28.4 & 30.9 & 31.4 & 31.0 & 31.8 & 33.9 & 29.6 & 31.0 
(\textcolor{MidnightBlue}{-20.9\%})\\
&\cc\textbf{HiKER-SGG} & \cc\textbf{32.7} & \cc\textbf{33.8} & \cc\textbf{32.6} & \cc\textbf{36.0} & \cc\textbf{30.4} & \cc\textbf{35.7} & \cc\textbf{34.7} & \cc\textbf{36.3} & \cc\textbf{36.7} & \cc\textbf{39.9} & \cc\textbf{39.7} & \cc\textbf{31.1} & \cc\textbf{34.2} & \cc\textbf{32.7} & \cc\textbf{35.4} & \cc\textbf{34.9} & \cc\textbf{35.4} & \cc\textbf{37.1} & \cc\textbf{39.2} & \cc\textbf{32.6} & \cc\textbf{35.1} 
 (\textcolor{MidnightBlue}{\cc\textbf{-14.8\%}})\\
\bottomrule
\end{tabular}
}
\vspace{-20pt}
\end{center}
\end{table*}
\subsection{Experimental Settings}
\label{sec:expsetting}
\looseness=-1
\textbf{Datasets.} 
Following the literature~\cite{zareian2020bridging,chen2023more}, we conduct experiments using the widely recognized Visual Genome (VG)~\cite{krishna2017visual} dataset, which consists of 108,077 images, each annotated with objects and relations. Following previous work~\cite{xu2017scene}, we filter the dataset to use the most frequent 150 object classes and 50 predicate classes for experiments. 

To standardize and evaluate SGG robustness, we create a \textbf{corrupted Visual Genome (VG-C) benchmark}, which comprises 20 corruption types designed to simulate realistic corruptions that may occur in real-world scenarios. Specifically, the first 15 types of corruption introduced by Hendrycks \textit{et al.}~\cite{hendrycks2018benchmarking} are widely recognized as standard benchmarks for evaluating robustness. To further align with real-world scenarios, we introduce 5 additional types of \textit{natural} corruption\footnote{Here, \textit{natural} corruptions refer to image degradations that arise from real-world environmental factors affecting the scene being captured.} to our evaluation: sun glare, water-drop, wildfire smoke, rain, and dust. 
A detailed description and visualization of the VG-C dataset are provided in Section B.2 of the Supplementary Materials.

\textbf{Tasks and Metrics}. 
We assess the effectiveness of our proposed approach in the context of two standard SGG tasks: Predicate Classification (PredCls) and Scene Graph Classification (SGCls). We evaluate the performance of the SGG models by top-$k$ mean triplet recall (mR@$k$) metric on both the PredCls and SGCls tasks. We also report the constrained (C) and unconstrained (UC) performance results, depending on the presence or absence of the graph constraint. This constraint restricts our SGG model to predict only a single relation between each pair of objects.

\looseness=-1
\textbf{Implementation Details.} We use the Faster-RCNN~\cite{ren2015faster} as the object detector, which is based on VGG-16~\cite{simonyan2015very} backbone provided by Zellers \textit{et al.}~\cite{zellers2018neural}. Regarding \ttt{FCNet} in Equations (\ref{fcn1}) and (\ref{fcn2}), we follow GB-Net~\cite{zareian2020bridging} to use 3-layer fully connected networks with \ttt{ReLU} activation. We set the message propagation steps $t=3$ and use a 1024-dimensional vector to represent each node. The hyperparameter $\alpha$ in Equation (\ref{eq:update}) is set to 0.9.
We also adopt the BPL~\cite{guo2021general} method to train our SGG model with unbiased data. 
In our experiments, we train our model for 30 epochs, initializing the learning rate at $1\times 10^{-4}$. 
A single NVIDIA Quadro RTX 6000 GPU is used for all the experiments.

\textbf{Baselines.} We compare our performance with the following state-of-the-art SGG methods: IMP+~\cite{xu2017scene}, Neural Motifs~\cite{zellers2018neural}, VCTree~\cite{tang2019learning}, PCPL~\cite{yan2020pcpl}, CogTree~\cite{yu2020cogtree}, EBM~\cite{suhail2021energy}, G2S~\cite{guo2021general}, DLFE~\cite{chiou2021recovering}, RTPB~\cite{chen2022resistance}, PPDL~\cite{li2022ppdl}, NICE~\cite{li2022devil}, NARE~\cite{goel2022not}, HML~\cite{deng2022hierarchical}, SQUAT~\cite{jung2023devil}, PE-Net~\cite{zheng2023prototype}, PE-Net + SIL~\cite{wang2023improving}. Additionally, we compare our approach with SGG methods that are knowledge graph-based, which are closely related to our work: GB-Net~\cite{zareian2020bridging} and EB-Net + EOA~\cite{chen2023more}. 
For a fair comparison, we present the performance results of these methods directly from their respective original papers.

\begin{figure*}[t]
\vspace{-2pt}
\centering
\includegraphics[width=\textwidth]{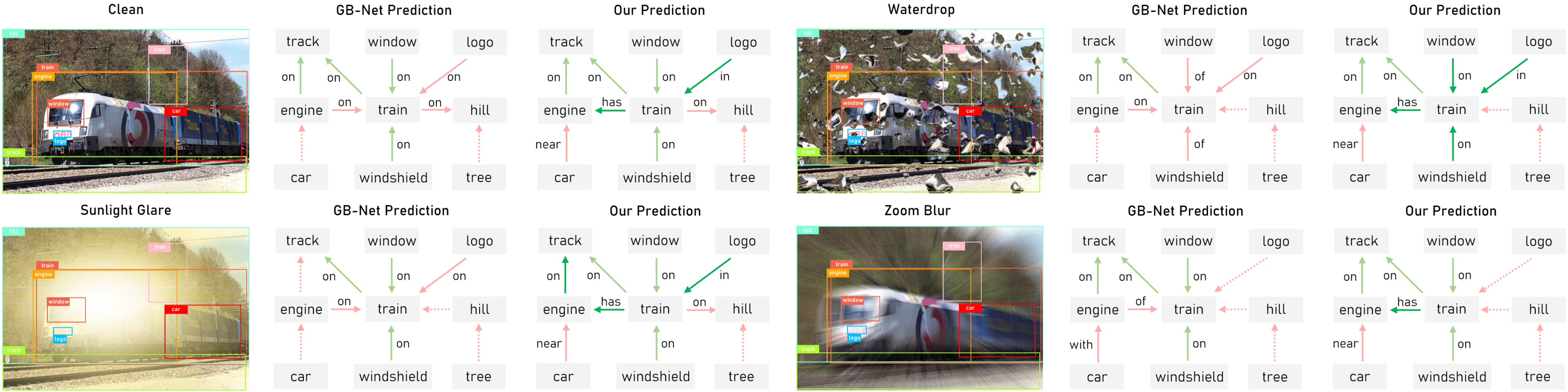}
\vspace{-16pt}
\caption{\textbf{Qualitative comparisons on the PredCls task}.  The visualized predicted predicates are picked from the top 50 predicted triplets. Here, \textcolor[RGB]{255,171,171}{red} dashed lines denote undetected predicates, solid \textcolor[RGB]{255,171,171}{red} lines denote incorrect predictions, and solid \textcolor[RGB]{169,209,142}{green} lines indicate correct predictions. For an easier comparison, predicates correctly predicted by our method but incorrectly by GB-Net are highlighted in \textcolor[RGB]{0,176,80}{dark green}.}
\label{fig:qualitative}
\vspace{-14pt}
\end{figure*}

\subsection{Results and Discussions}
\textbf{Quantitative Results.} In Table \ref{tab:performance}, we report our performance results for the PredCls task and SGCls tasks on clean images in the Visual Genome~\cite{krishna2017visual} dataset. With the hierarchical predicate prediction paradigm, our method consistently outperforms the knowledge graph-based GB-Net~\cite{zareian2020bridging} and EB-Net + EOA~\cite{chen2023more} methods. When compared with other state-of-the-art SGG methods, our HiKER-SGG still achieves competitive performance in terms of mean recall. 

\looseness=-1
We also show our results on the VG-C dataset in Table \ref{tab:performance2} to demonstrate our method also generalizes well to unseen real-world corruptions.  Specifically, we compare our performance with that of the knowledge graph-based methods across all six metrics. 
Table \ref{tab:performance2} illustrates that our method achieves an average improvement of around 4\% across all six metrics for all 20 types of corruption. Moreover, relative to the clean image benchmark, our method exhibits a lower percentage of performance degradation, showcasing our model's resilience in handling such corrupted scenarios. For instance, in the presence of impulse noise corruption, our mR@20, when considering graph constraints, experiences an 8.6\% reduction, dropping from 33.4\% to 24.8\%. In comparison, the EB-Net~\cite{chen2023more} method shows a greater 10.4\% degradation, decreasing from 30.8\% to 20.4\%.

\looseness=-1
\textbf{Qualitative Results.} To provide further insights into the effectiveness of our method, we visualize some scene graphs generated by our method and the baseline GB-Net~\cite{zareian2020bridging} method, under both clean and corrupted scenarios in Figure \ref{fig:qualitative}. In the upper left section of the image, we can observe the scene graphs generated by both methods on the clean image. Notably, while GB-Net tends to predict more general predicate classes (\textit{e.g.}, \ttt{on}), our method accurately predicts the $\langle$\ttt{train}-\ttt{has}-\ttt{engine}$\rangle$ and $\langle$\ttt{logo}-\ttt{in}-\ttt{train}$\rangle$ triplets.

\looseness=-1
We also illustrate the SGG results under sun glare, water-drop, and zoom blur corruptions obtained by both methods in Figure \ref{fig:qualitative}.
In these challenging scenarios, non-hierarchical GB-Net~\cite{zareian2020bridging}, struggles to detect the relation since the region feature is corrupted. In comparison, our method firstly determines the superclass relation rather than directly proceeding to subclass classification.
This strategy enhances the robustness of our proposed method, enabling it to consistently generate a similar scene graph as in clean images. 

\begin{table}[t]
\renewcommand\arraystretch{0.9}
\centering
\caption{\textbf{Ablation studies on the PredCls task using VG dataset}. PH and EH refer to predicate and entity hierarchical prediction heads respectively, and $\mathcal{M}$/$\mathcal{D}$ indicate whether these hierarchies are manually configured ($\mathcal{M}$) following Zellers \textit{et al.}~\cite{zellers2018neural} or discovered ($\mathcal{D}$) by hierarchical clustering. AR refers to adaptive refinement.}
\vspace{-7pt}
\label{tab:ablation}
\resizebox{\linewidth}{!}{
\begin{tabular}{ccc|ccc}
\toprule
PH & EH & AR  & mR@20: UC/C & mR@50: UC/C & mR@100: UC/C \\ \midrule
\XSolidBrush & \XSolidBrush &  \XSolidBrush   &  39.8 / 30.8 & 54.9 / 36.7 & 66.3 / 39.2 
 \\
\XSolidBrush & \XSolidBrush &  \Checkmark   &  40.4 / 31.4 & 55.7 / 37.2 & 67.1 / 39.8 
 \\
$\mathcal{M}$ & \XSolidBrush &  \XSolidBrush &   41.6 / 32.9 & 57.3 / 37.5 & 68.1 / 39.6
   \\
$\mathcal{M}$ & $\mathcal{M}$ &  \XSolidBrush &   41.4 / 33.1 & 57.6 / 37.9 & 68.2 / 39.7 
 \\
$\mathcal{M}$ &$\mathcal{M}$  & \Checkmark & 41.8 / 33.2 & 57.7 / 38.1 & 68.7 / 40.0
\\ 
$\mathcal{D}$& $\mathcal{D}$& \XSolidBrush &  41.7 / 33.2 & 57.7 / 38.8 & 69.0 / 40.4
\\ 
 \rowcolor{gray!20}
$\mathcal{D}$ &$\mathcal{D}$  & \Checkmark & \textbf{42.1} / \textbf{33.4} & \textbf{57.9} / \textbf{39.3} & \textbf{69.2} / \textbf{41.2}
\\ 
\bottomrule
\end{tabular}
}
\vspace{-15pt}
\end{table}

\subsection{Ablation Studies}
\label{sec:ablation}
\textbf{Effectiveness of Each Component}. To systematically analyze the impacts of different components in HiKER-SGG, we conduct an ablation study on the Visual Genome~\cite{krishna2017visual} dataset in Table \ref{tab:ablation}. We have the following key observations: (1) The inclusion of the hierarchical inference process for predicate alone enhances the mR@$k$ by 1.0\%, and adding the hierarchical inference process for entity further boosts mR@$k$ by an additional 0.5\%; (2) Replacing manually configured hierarchical structures with those discovered ones yields a non-trivial 0.4\%$\sim$0.7\% increase in mR@$k$; (3) Implementing the adaptive refinement contributes to a further improvement in performance by 0.2\%$\sim$0.8\% mR@$k$.

\textbf{Hyperparameter Analysis for $\alpha$}. We conduct experiments with five distinct values for the hyperparameter $\alpha$ and report the mR under the PredCls setting in Table \ref{tab:hyper}. We can observe that our setting of $\alpha=0.9$ yields the highest performance. The reason may be that this optimal value effectively balances the surface-level and deeper biases among the predicate and entity classes, which contributes to the improved unbiased prediction capabilities of our HiKER-SGG model.

\begin{figure}[t]
\begin{minipage}[t]{0.48\linewidth}
\makeatletter\def\@captype{table}
\caption{Hyperparameter analysis for $\alpha$ in Equation (\ref{eq:update}).}
\vspace{-6.5pt}
\label{tab:hyper}
\resizebox{\linewidth}{!}{
\begin{tabular}{lcc}
\toprule
Value of $\alpha$ & mR@50      & mR@100  \\ \midrule
$\alpha=0.5$ & 56.7 / 38.1  & 66.9 / 40.0 \\ 
$\alpha=0.8$ & 57.4 / 38.5 & 68.5 / 40.7 \\
$\cc \alpha=0.9$ & \cc \textbf{57.9} / \textbf{39.3} & \cc \textbf{69.2} / \textbf{41.2} \\
$\alpha=0.95$ & 57.6 / 38.9 & 69.1 / 40.9 \\
$\alpha=0.99$ & 57.6 / 38.7 & 68.8 / 40.5 \\
\bottomrule
\end{tabular}
}
\end{minipage}
\hspace{2pt}
\begin{minipage}[t]{0.48\linewidth}
\makeatletter\def\@captype{table}
\caption{Training time and parameter count of HiKER-SGG compared with other methods.}
\vspace{-8pt}
\label{tab:efficiency}
\resizebox{\linewidth}{!}{
\begin{tabular}{lcc}
\toprule
Method & Training      & \# params \\ \midrule
KERN~\cite{chen2019knowledge} & 179.1 min & 405.2M \\
GB-Net~\cite{zareian2020bridging} & 84.6 min & 444.6M \\
EB-Net~\cite{chen2023more} & 89.7 min & 448.8M \\
\cc \textbf{HiKER-SGG} & \cc 101.3 min & \cc 455.9M \\
\bottomrule
\end{tabular}
}
\end{minipage}
\vspace{-15pt}
\end{figure}

\textbf{Efficiency Comparison}. We also compare the training time and the number of parameters of our HiKER-SGG with other methods in Table \ref{tab:efficiency}. Our HiKER-SGG divides a general classifier into multiple smaller hierarchical classifiers, thereby maintaining relatively high efficiency compared to non-hierarchical methods such as GB-Net~\cite{zareian2020bridging} and EB-Net~\cite{chen2023more}. Specifically, while incorporating only 7M additional parameters and extending the training time by only 12 minutes per epoch, our HiKER-SGG exhibits significantly enhanced robustness with both clean and corrupted images.

\section{Conclusion}
\looseness=-1
In this work, we first introduce a novel task, robust SGG in the presence of real-world corruptions. To address the challenge of interpreting visual scenes with corruptions, we then propose the \textbf{Hi}erarchical \textbf{K}nowledge \textbf{E}nhanced \textbf{R}obust \textbf{S}cene \textbf{G}raph \textbf{G}eneration (HiKER-SGG) framework. HiKER-SGG is corruption-agnostic, trained exclusively on clean images yet tested on corrupted ones without further training. It leverages hierarchical knowledge from external sources and a hierarchical prediction head, serving as an algorithmic prior for decision-making, to effectively reason and correct inaccuracies.
Moreover, we developed a corrupted Visual Genome (VG-C) benchmark with 20 different corruptions to standardize and evaluate SGG robustness. 
Through extensive experiments, we have demonstrated that HiKER-SGG outperforms the state-of-the-art models on both clean and corrupted images.

\section*{Acknowledgement}
This work has been funded in part by the Army Research Laboratory (ARL) under grant W911NF-23-2-0007 and W911NF-19-2-0146, and the
Air Force Office of Scientific Research (AFOSR) under grants FA9550-18-1-0097 and FA9550-18-1-0251.

{
\small
\bibliographystyle{ieeenat_fullname}
\bibliography{main}

\begin{thebibliography}{93}
\providecommand{\natexlab}[1]{#1}
\providecommand{\url}[1]{\texttt{#1}}
\expandafter\ifx\csname urlstyle\endcsname\relax
  \providecommand{\doi}[1]{doi: #1}\else
  \providecommand{\doi}{doi: \begingroup \urlstyle{rm}\Url}\fi

\bibitem[Agarwal et~al.(2020)Agarwal, Mangal, et~al.]{agarwal2020visual}
Aniket Agarwal, Ayush Mangal, et~al.
\newblock Visual relationship detection using scene graphs: A survey.
\newblock \emph{arXiv preprint arXiv:2005.08045}, 2020.

\bibitem[Bhagat et~al.(2023{\natexlab{a}})Bhagat, Stepputtis, Campbell, and Sycara]{bhagat2023knowledge}
Sarthak Bhagat, Simon Stepputtis, Joseph Campbell, and Katia Sycara.
\newblock Knowledge-guided short-context action anticipation in human-centric videos.
\newblock \emph{arXiv preprint arXiv:2309.05943}, 2023{\natexlab{a}}.

\bibitem[Bhagat et~al.(2023{\natexlab{b}})Bhagat, Stepputtis, Campbell, and Sycara]{bhagat2023sample}
Sarthak Bhagat, Simon Stepputtis, Joseph Campbell, and Katia Sycara.
\newblock Sample-efficient learning of novel visual concepts.
\newblock In \emph{CoLLAs}, pages 637--657. PMLR, 2023{\natexlab{b}}.

\bibitem[Caron et~al.(2021)Caron, Touvron, Misra, J{\'e}gou, Mairal, Bojanowski, and Joulin]{caron2021emerging}
Mathilde Caron, Hugo Touvron, Ishan Misra, Herv{\'e} J{\'e}gou, Julien Mairal, Piotr Bojanowski, and Armand Joulin.
\newblock Emerging properties in self-supervised vision transformers.
\newblock In \emph{ICCV}, pages 9650--9660, 2021.

\bibitem[Chang et~al.(2021)Chang, Ren, Xu, Li, Chen, and Hauptmann]{chang2021comprehensive}
Xiaojun Chang, Pengzhen Ren, Pengfei Xu, Zhihui Li, Xiaojiang Chen, and Alex Hauptmann.
\newblock A comprehensive survey of scene graphs: Generation and application.
\newblock \emph{IEEE TPAMI}, 45\penalty0 (1):\penalty0 1--26, 2021.

\bibitem[Chen et~al.(2022)Chen, Zhan, Yu, Liu, Luo, and Du]{chen2022resistance}
Chao Chen, Yibing Zhan, Baosheng Yu, Liu Liu, Yong Luo, and Bo Du.
\newblock Resistance training using prior bias: toward unbiased scene graph generation.
\newblock In \emph{AAAI}, pages 212--220, 2022.

\bibitem[Chen et~al.(2019{\natexlab{a}})Chen, Liu, Liu, and Wang]{chen2019towards}
Po-Yi Chen, Alexander~H Liu, Yen-Cheng Liu, and Yu-Chiang~Frank Wang.
\newblock Towards scene understanding: Unsupervised monocular depth estimation with semantic-aware representation.
\newblock In \emph{CVPR}, pages 2624--2632, 2019{\natexlab{a}}.

\bibitem[Chen et~al.(2019{\natexlab{b}})Chen, Yu, Chen, and Lin]{chen2019knowledge}
Tianshui Chen, Weihao Yu, Riquan Chen, and Liang Lin.
\newblock Knowledge-embedded routing network for scene graph generation.
\newblock In \emph{CVPR}, pages 6163--6171, 2019{\natexlab{b}}.

\bibitem[Chen et~al.(2023)Chen, Rezayi, and Li]{chen2023more}
Zhanwen Chen, Saed Rezayi, and Sheng Li.
\newblock More knowledge, less bias: Unbiasing scene graph generation with explicit ontological adjustment.
\newblock In \emph{WACV}, pages 4023--4032, 2023.

\bibitem[Chiou et~al.(2021)Chiou, Ding, Yan, Wang, Zimmermann, and Feng]{chiou2021recovering}
Meng-Jiun Chiou, Henghui Ding, Hanshu Yan, Changhu Wang, Roger Zimmermann, and Jiashi Feng.
\newblock Recovering the unbiased scene graphs from the biased ones.
\newblock In \emph{ACM MM}, pages 1581--1590, 2021.

\bibitem[Cho et~al.(2014)Cho, van Merri{\"e}nboer, Gul{\c{c}}ehre, Bahdanau, Bougares, Schwenk, and Bengio]{cho2014learning}
Kyunghyun Cho, Bart van Merri{\"e}nboer, {\c{C}}a{\u{g}}lar Gul{\c{c}}ehre, Dzmitry Bahdanau, Fethi Bougares, Holger Schwenk, and Yoshua Bengio.
\newblock Learning phrase representations using rnn encoder--decoder for statistical machine translation.
\newblock In \emph{EMNLP}, pages 1724--1734, 2014.

\bibitem[Dai et~al.(2017)Dai, Zhang, and Lin]{dai2017detecting}
Bo Dai, Yuqi Zhang, and Dahua Lin.
\newblock Detecting visual relationships with deep relational networks.
\newblock In \emph{CVPR}, pages 3076--3086, 2017.

\bibitem[Deng et~al.(2022)Deng, Li, Zhang, Xiang, Wang, Chen, and Ma]{deng2022hierarchical}
Youming Deng, Yansheng Li, Yongjun Zhang, Xiang Xiang, Jian Wang, Jingdong Chen, and Jiayi Ma.
\newblock Hierarchical memory learning for fine-grained scene graph generation.
\newblock In \emph{ECCV}, pages 266--283. Springer, 2022.

\bibitem[Desai et~al.(2021)Desai, Wu, Tripathi, and Vasconcelos]{desai2021learning}
Alakh Desai, Tz-Ying Wu, Subarna Tripathi, and Nuno Vasconcelos.
\newblock Learning of visual relations: The devil is in the tails.
\newblock In \emph{ICCV}, pages 15404--15413, 2021.

\bibitem[Dong et~al.(2022)Dong, Gan, Song, Wu, Cheng, and Nie]{dong2022stacked}
Xingning Dong, Tian Gan, Xuemeng Song, Jianlong Wu, Yuan Cheng, and Liqiang Nie.
\newblock Stacked hybrid-attention and group collaborative learning for unbiased scene graph generation.
\newblock In \emph{CVPR}, pages 19427--19436, 2022.

\bibitem[Eslami et~al.(2016)Eslami, Heess, Weber, Tassa, Szepesvari, Hinton, et~al.]{eslami2016attend}
SM Eslami, Nicolas Heess, Theophane Weber, Yuval Tassa, David Szepesvari, Geoffrey~E Hinton, et~al.
\newblock Attend, infer, repeat: Fast scene understanding with generative models.
\newblock In \emph{NeurIPS}, 2016.

\bibitem[Fang et~al.(2023)Fang, Dong, Bao, Wang, and Wei]{fang2023corrupted}
Yuxin Fang, Li Dong, Hangbo Bao, Xinggang Wang, and Furu Wei.
\newblock Corrupted image modeling for self-supervised visual pre-training.
\newblock In \emph{ICLR}, 2023.

\bibitem[Farshad et~al.(2023)Farshad, Yeganeh, Chi, Shen, Ommer, and Navab]{farshad2023scenegenie}
Azade Farshad, Yousef Yeganeh, Yu Chi, Chengzhi Shen, B{\"o}jrn Ommer, and Nassir Navab.
\newblock Scenegenie: Scene graph guided diffusion models for image synthesis.
\newblock In \emph{ICCV}, pages 88--98, 2023.

\bibitem[Goel et~al.(2022)Goel, Fernando, Keller, and Bilen]{goel2022not}
Arushi Goel, Basura Fernando, Frank Keller, and Hakan Bilen.
\newblock Not all relations are equal: Mining informative labels for scene graph generation.
\newblock In \emph{CVPR}, pages 15596--15606, 2022.

\bibitem[Gray et~al.(2023)Gray, Moraes, Bian, Wang, Tian, Wilson, Huang, Xiong, and Guo]{gray2023glare}
Nicholas Gray, Megan Moraes, Jiang Bian, Alex Wang, Allen Tian, Kurt Wilson, Yan Huang, Haoyi Xiong, and Zhishan Guo.
\newblock Glare: A dataset for traffic sign detection in sun glare.
\newblock \emph{IEEE TITS}, 2023.

\bibitem[Gu et~al.(2019)Gu, Zhao, Lin, Li, Cai, and Ling]{gu2019scene}
Jiuxiang Gu, Handong Zhao, Zhe Lin, Sheng Li, Jianfei Cai, and Mingyang Ling.
\newblock Scene graph generation with external knowledge and image reconstruction.
\newblock In \emph{CVPR}, pages 1969--1978, 2019.

\bibitem[Guo et~al.(2021)Guo, Gao, Wang, Hu, Xu, Lu, Shen, and Song]{guo2021general}
Yuyu Guo, Lianli Gao, Xuanhan Wang, Yuxuan Hu, Xing Xu, Xu Lu, Heng~Tao Shen, and Jingkuan Song.
\newblock From general to specific: Informative scene graph generation via balance adjustment.
\newblock In \emph{ICCV}, pages 16383--16392, 2021.

\bibitem[Halder et~al.(2019)Halder, Lalonde, and Charette]{halder2019physics}
Shirsendu~Sukanta Halder, Jean-Fran{\c{c}}ois Lalonde, and Raoul~de Charette.
\newblock Physics-based rendering for improving robustness to rain.
\newblock In \emph{ICCV}, pages 10203--10212, 2019.

\bibitem[Han et~al.(2022)Han, Dong, Song, Gan, Zhan, Yan, and Nie]{han2022divide}
Xianjing Han, Xingning Dong, Xuemeng Song, Tian Gan, Yibing Zhan, Yan Yan, and Liqiang Nie.
\newblock Divide-and-conquer predictor for unbiased scene graph generation.
\newblock \emph{IEEE TCSVT}, 32\penalty0 (12):\penalty0 8611--8622, 2022.

\bibitem[He et~al.(2016)He, Zhang, Ren, and Sun]{he2016deep}
Kaiming He, Xiangyu Zhang, Shaoqing Ren, and Jian Sun.
\newblock Deep residual learning for image recognition.
\newblock In \emph{CVPR}, pages 770--778, 2016.

\bibitem[He et~al.(2021)He, Gao, Song, Cai, and Li]{he2021learning}
Tao He, Lianli Gao, Jingkuan Song, Jianfei Cai, and Yuan-Fang Li.
\newblock Learning from the scene and borrowing from the rich: tackling the long tail in scene graph generation.
\newblock In \emph{IJCAI}, pages 587--593, 2021.

\bibitem[He et~al.(2022)He, Gao, Song, and Li]{he2022state}
Tao He, Lianli Gao, Jingkuan Song, and Yuan-Fang Li.
\newblock State-aware compositional learning toward unbiased training for scene graph generation.
\newblock \emph{IEEE TIP}, 32:\penalty0 43--56, 2022.

\bibitem[Hendrycks and Dietterich(2018)]{hendrycks2018benchmarking}
Dan Hendrycks and Thomas Dietterich.
\newblock Benchmarking neural network robustness to common corruptions and perturbations.
\newblock In \emph{ICLR}, 2018.

\bibitem[Hendrycks et~al.(2019)Hendrycks, Mu, Cubuk, Zoph, Gilmer, and Lakshminarayanan]{hendrycks2019augmix}
Dan Hendrycks, Norman Mu, Ekin~Dogus Cubuk, Barret Zoph, Justin Gilmer, and Balaji Lakshminarayanan.
\newblock Augmix: A simple data processing method to improve robustness and uncertainty.
\newblock In \emph{ICLR}, 2019.

\bibitem[Herrmann et~al.(2022)Herrmann, Sargent, Jiang, Zabih, Chang, Liu, Krishnan, and Sun]{herrmann2022pyramid}
Charles Herrmann, Kyle Sargent, Lu Jiang, Ramin Zabih, Huiwen Chang, Ce Liu, Dilip Krishnan, and Deqing Sun.
\newblock Pyramid adversarial training improves vit performance.
\newblock In \emph{CVPR}, pages 13419--13429, 2022.

\bibitem[Jia et~al.(2023)Jia, Ding, Pang, Gao, Xin, Hu, and Nie]{jia2023image}
Junhua Jia, Xiangqian Ding, Shunpeng Pang, Xiaoyan Gao, Xiaowei Xin, Ruotong Hu, and Jie Nie.
\newblock Image captioning based on scene graphs: A survey.
\newblock \emph{Expert Systems with Applications}, page 120698, 2023.

\bibitem[Jiang and Taylor(2023)]{jiang2023scene}
Bowen Jiang and Camillo~J Taylor.
\newblock Scene graph generation from hierarchical relationship reasoning.
\newblock \emph{arXiv preprint arXiv:2303.06842}, 2023.

\bibitem[Johnson et~al.(2015)Johnson, Krishna, Stark, Li, Shamma, Bernstein, and Fei-Fei]{johnson2015image}
Justin Johnson, Ranjay Krishna, Michael Stark, Li-Jia Li, David Shamma, Michael Bernstein, and Li Fei-Fei.
\newblock Image retrieval using scene graphs.
\newblock In \emph{CVPR}, pages 3668--3678, 2015.

\bibitem[Johnson et~al.(2018)Johnson, Gupta, and Fei-Fei]{johnson2018image}
Justin Johnson, Agrim Gupta, and Li Fei-Fei.
\newblock Image generation from scene graphs.
\newblock In \emph{CVPR}, pages 1219--1228, 2018.

\bibitem[Johnson(1967)]{johnson1967hierarchical}
Stephen~C Johnson.
\newblock Hierarchical clustering schemes.
\newblock \emph{Psychometrika}, 32\penalty0 (3):\penalty0 241--254, 1967.

\bibitem[Jung et~al.(2023)Jung, Kim, Kim, and Cho]{jung2023devil}
Deunsol Jung, Sanghyun Kim, Won~Hwa Kim, and Minsu Cho.
\newblock Devil's on the edges: Selective quad attention for scene graph generation.
\newblock In \emph{CVPR}, pages 18664--18674, 2023.

\bibitem[Kireev et~al.(2022)Kireev, Andriushchenko, and Flammarion]{kireev2022effectiveness}
Klim Kireev, Maksym Andriushchenko, and Nicolas Flammarion.
\newblock On the effectiveness of adversarial training against common corruptions.
\newblock In \emph{UAI}, pages 1012--1021. PMLR, 2022.

\bibitem[Krishna et~al.(2017)Krishna, Zhu, Groth, Johnson, Hata, Kravitz, Chen, Kalantidis, Li, Shamma, et~al.]{krishna2017visual}
Ranjay Krishna, Yuke Zhu, Oliver Groth, Justin Johnson, Kenji Hata, Joshua Kravitz, Stephanie Chen, Yannis Kalantidis, Li-Jia Li, David~A Shamma, et~al.
\newblock Visual genome: Connecting language and vision using crowdsourced dense image annotations.
\newblock \emph{IJCV}, 123:\penalty0 32--73, 2017.

\bibitem[Lei et~al.(2023)Lei, Gao, Wu, Wang, Liu, Zhang, and Shou]{lei2023symbolic}
Stan~Weixian Lei, Difei Gao, Jay~Zhangjie Wu, Yuxuan Wang, Wei Liu, Mengmi Zhang, and Mike~Zheng Shou.
\newblock Symbolic replay: Scene graph as prompt for continual learning on vqa task.
\newblock In \emph{AAAI}, pages 1250--1259, 2023.

\bibitem[Li et~al.(2022{\natexlab{a}})Li, Chen, Huang, Zhang, Zhang, and Xiao]{li2022devil}
Lin Li, Long Chen, Yifeng Huang, Zhimeng Zhang, Songyang Zhang, and Jun Xiao.
\newblock The devil is in the labels: Noisy label correction for robust scene graph generation.
\newblock In \emph{CVPR}, pages 18869--18878, 2022{\natexlab{a}}.

\bibitem[Li et~al.(2023)Li, Xiao, Shi, Wang, Shao, Liu, Yang, and Chen]{li2023label}
Lin Li, Jun Xiao, Hanrong Shi, Wenxiao Wang, Jian Shao, An-An Liu, Yi Yang, and Long Chen.
\newblock Label semantic knowledge distillation for unbiased scene graph generation.
\newblock \emph{IEEE TCSVT}, 2023.

\bibitem[Li et~al.(2021)Li, Zhang, Wan, and He]{li2021bipartite}
Rongjie Li, Songyang Zhang, Bo Wan, and Xuming He.
\newblock Bipartite graph network with adaptive message passing for unbiased scene graph generation.
\newblock In \emph{CVPR}, pages 11109--11119, 2021.

\bibitem[Li et~al.(2022{\natexlab{b}})Li, Zhang, Bai, Zhao, Jiang, and Yuan]{li2022ppdl}
Wei Li, Haiwei Zhang, Qijie Bai, Guoqing Zhao, Ning Jiang, and Xiaojie Yuan.
\newblock Ppdl: Predicate probability distribution based loss for unbiased scene graph generation.
\newblock In \emph{CVPR}, pages 19447--19456, 2022{\natexlab{b}}.

\bibitem[Li and Jiang(2019)]{li2019know}
Xiangyang Li and Shuqiang Jiang.
\newblock Know more say less: Image captioning based on scene graphs.
\newblock \emph{IEEE TMM}, 21\penalty0 (8):\penalty0 2117--2130, 2019.

\bibitem[Li et~al.(2022{\natexlab{c}})Li, Chen, Shao, Xiao, Zhang, and Xiao]{li2022rethinking}
Xingchen Li, Long Chen, Jian Shao, Shaoning Xiao, Songyang Zhang, and Jun Xiao.
\newblock Rethinking the evaluation of unbiased scene graph generation.
\newblock In \emph{BMVC}, 2022{\natexlab{c}}.

\bibitem[Li et~al.(2022{\natexlab{d}})Li, Guo, Liu, and Sun]{li2022embodied}
Xinghang Li, Di Guo, Huaping Liu, and Fuchun Sun.
\newblock Embodied semantic scene graph generation.
\newblock In \emph{CoRL}, pages 1585--1594. PMLR, 2022{\natexlab{d}}.

\bibitem[Li et~al.(2016)Li, Zemel, Brockschmidt, and Tarlow]{li2016gated}
Yujia Li, Richard Zemel, Marc Brockschmidt, and Daniel Tarlow.
\newblock Gated graph sequence neural networks.
\newblock In \emph{ICLR}, 2016.

\bibitem[Li et~al.(2018)Li, Ouyang, Zhou, Shi, Zhang, and Wang]{li2018factorizable}
Yikang Li, Wanli Ouyang, Bolei Zhou, Jianping Shi, Chao Zhang, and Xiaogang Wang.
\newblock Factorizable net: an efficient subgraph-based framework for scene graph generation.
\newblock In \emph{ECCV}, pages 335--351. Springer, 2018.

\bibitem[Miller(1995)]{miller1995wordnet}
George~A Miller.
\newblock Wordnet: a lexical database for english.
\newblock \emph{Communications of the ACM}, 38\penalty0 (11):\penalty0 39--41, 1995.

\bibitem[Mintun et~al.(2021)Mintun, Kirillov, and Xie]{mintun2021interaction}
Eric Mintun, Alexander Kirillov, and Saining Xie.
\newblock On interaction between augmentations and corruptions in natural corruption robustness.
\newblock In \emph{NeurIPS}, pages 3571--3583, 2021.

\bibitem[Mirza et~al.(2022)Mirza, Micorek, Possegger, and Bischof]{mirza2022norm}
M~Jehanzeb Mirza, Jakub Micorek, Horst Possegger, and Horst Bischof.
\newblock The norm must go on: Dynamic unsupervised domain adaptation by normalization.
\newblock In \emph{CVPR}, pages 14765--14775, 2022.

\bibitem[Pennington et~al.(2014)Pennington, Socher, and Manning]{pennington2014glove}
Jeffrey Pennington, Richard Socher, and Christopher~D Manning.
\newblock Glove: Global vectors for word representation.
\newblock In \emph{EMNLP}, pages 1532--1543, 2014.

\bibitem[Qian et~al.(2023)Qian, Chen, Chen, Wu, and Jiang]{qian2022scene}
Tianwen Qian, Jingjing Chen, Shaoxiang Chen, Bo Wu, and Yu-Gang Jiang.
\newblock Scene graph refinement network for visual question answering.
\newblock \emph{IEEE TMM}, 25:\penalty0 3950--3961, 2023.

\bibitem[Quan et~al.(2019)Quan, Deng, Chen, and Ji]{quan2019deep}
Yuhui Quan, Shijie Deng, Yixin Chen, and Hui Ji.
\newblock Deep learning for seeing through window with raindrops.
\newblock In \emph{ICCV}, pages 2463--2471, 2019.

\bibitem[Radford et~al.(2021)Radford, Kim, Hallacy, Ramesh, Goh, Agarwal, Sastry, Askell, Mishkin, Clark, et~al.]{radford2021learning}
Alec Radford, Jong~Wook Kim, Chris Hallacy, Aditya Ramesh, Gabriel Goh, Sandhini Agarwal, Girish Sastry, Amanda Askell, Pamela Mishkin, Jack Clark, et~al.
\newblock Learning transferable visual models from natural language supervision.
\newblock In \emph{ICML}, pages 8748--8763. PMLR, 2021.

\bibitem[Ren et~al.(2015)Ren, He, Girshick, and Sun]{ren2015faster}
Shaoqing Ren, Kaiming He, Ross Girshick, and Jian Sun.
\newblock Faster r-cnn: Towards real-time object detection with region proposal networks.
\newblock In \emph{NeurIPS}, page 91–99, 2015.

\bibitem[Rusak et~al.(2020)Rusak, Schott, Zimmermann, Bitterwolf, Bringmann, Bethge, and Brendel]{rusak2020simple}
Evgenia Rusak, Lukas Schott, Roland~S Zimmermann, Julian Bitterwolf, Oliver Bringmann, Matthias Bethge, and Wieland Brendel.
\newblock A simple way to make neural networks robust against diverse image corruptions.
\newblock In \emph{ECCV}, pages 53--69. Springer, 2020.

\bibitem[Simonyan and Zisserman(2015)]{simonyan2015very}
Karen Simonyan and Andrew Zisserman.
\newblock Very deep convolutional networks for large-scale image recognition.
\newblock In \emph{ICLR}, 2015.

\bibitem[Speer et~al.(2017)Speer, Chin, and Havasi]{speer2017conceptnet}
Robyn Speer, Joshua Chin, and Catherine Havasi.
\newblock Conceptnet 5.5: An open multilingual graph of general knowledge.
\newblock In \emph{AAAI}, page 4444–4451, 2017.

\bibitem[Suhail et~al.(2021)Suhail, Mittal, Siddiquie, Broaddus, Eledath, Medioni, and Sigal]{suhail2021energy}
Mohammed Suhail, Abhay Mittal, Behjat Siddiquie, Chris Broaddus, Jayan Eledath, Gerard Medioni, and Leonid Sigal.
\newblock Energy-based learning for scene graph generation.
\newblock In \emph{CVPR}, pages 13936--13945, 2021.

\bibitem[Sun et~al.(2023)Sun, Zhi, Liao, Heikkil{\"a}, and Liu]{sun2023unbiased}
Shuzhou Sun, Shuaifeng Zhi, Qing Liao, Janne Heikkil{\"a}, and Li Liu.
\newblock Unbiased scene graph generation via two-stage causal modeling.
\newblock \emph{IEEE TPAMI}, 2023.

\bibitem[Tang et~al.(2019)Tang, Zhang, Wu, Luo, and Liu]{tang2019learning}
Kaihua Tang, Hanwang Zhang, Baoyuan Wu, Wenhan Luo, and Wei Liu.
\newblock Learning to compose dynamic tree structures for visual contexts.
\newblock In \emph{CVPR}, pages 6619--6628, 2019.

\bibitem[Tang et~al.(2020)Tang, Niu, Huang, Shi, and Zhang]{tang2020unbiased}
Kaihua Tang, Yulei Niu, Jianqiang Huang, Jiaxin Shi, and Hanwang Zhang.
\newblock Unbiased scene graph generation from biased training.
\newblock In \emph{CVPR}, pages 3716--3725, 2020.

\bibitem[Tang et~al.(2023{\natexlab{a}})Tang, Guo, and He]{tang2023cross}
Yushun Tang, Qinghai Guo, and Zhihai He.
\newblock Cross-inferential networks for source-free unsupervised domain adaptation.
\newblock In \emph{ICIP}, pages 96--100. IEEE, 2023{\natexlab{a}}.

\bibitem[Tang et~al.(2023{\natexlab{b}})Tang, Zhang, Xu, Chen, Cheng, Leng, Guo, and He]{tang2023neuro}
Yushun Tang, Ce Zhang, Heng Xu, Shuoshuo Chen, Jie Cheng, Luziwei Leng, Qinghai Guo, and Zhihai He.
\newblock Neuro-modulated hebbian learning for fully test-time adaptation.
\newblock In \emph{CVPR}, pages 3728--3738, 2023{\natexlab{b}}.

\bibitem[Tian et~al.(2021)Tian, Xu, Liu, Yan, Mao, Zhang, and Zhang]{tian2021mask}
Hongshuo Tian, Ning Xu, An-An Liu, Chenggang Yan, Zhendong Mao, Quan Zhang, and Yongdong Zhang.
\newblock Mask and predict: Multi-step reasoning for scene graph generation.
\newblock In \emph{ACM MM}, pages 4128--4136, 2021.

\bibitem[Tremblay et~al.(2021)Tremblay, Halder, De~Charette, and Lalonde]{tremblay2021rain}
Maxime Tremblay, Shirsendu~Sukanta Halder, Raoul De~Charette, and Jean-Fran{\c{c}}ois Lalonde.
\newblock Rain rendering for evaluating and improving robustness to bad weather.
\newblock \emph{IJCV}, 129:\penalty0 341--360, 2021.

\bibitem[Wald et~al.(2020)Wald, Dhamo, Navab, and Tombari]{wald2020learning}
Johanna Wald, Helisa Dhamo, Nassir Navab, and Federico Tombari.
\newblock Learning 3d semantic scene graphs from 3d indoor reconstructions.
\newblock In \emph{CVPR}, pages 3961--3970, 2020.

\bibitem[Wang et~al.(2023)Wang, Zhang, Huang, Ren, and Deng]{wang2023improving}
Jingyi Wang, Can Zhang, Jinfa Huang, Botao Ren, and Zhidong Deng.
\newblock Improving scene graph generation with superpixel-based interaction learning.
\newblock In \emph{ACM MM}, pages 1809--1820, 2023.

\bibitem[Wang et~al.(2020{\natexlab{a}})Wang, Wang, Yao, Shan, and Chen]{wang2020cross}
Sijin Wang, Ruiping Wang, Ziwei Yao, Shiguang Shan, and Xilin Chen.
\newblock Cross-modal scene graph matching for relationship-aware image-text retrieval.
\newblock In \emph{WACV}, pages 1508--1517, 2020{\natexlab{a}}.

\bibitem[Wang et~al.(2019)Wang, Wang, Shan, and Chen]{wang2019exploring}
Wenbin Wang, Ruiping Wang, Shiguang Shan, and Xilin Chen.
\newblock Exploring context and visual pattern of relationship for scene graph generation.
\newblock In \emph{CVPR}, pages 8188--8197, 2019.

\bibitem[Wang et~al.(2020{\natexlab{b}})Wang, Wang, Shan, and Chen]{wang2020sketching}
Wenbin Wang, Ruiping Wang, Shiguang Shan, and Xilin Chen.
\newblock Sketching image gist: Human-mimetic hierarchical scene graph generation.
\newblock In \emph{ECCV}, pages 222--239. Springer, 2020{\natexlab{b}}.

\bibitem[Wu et~al.(2023)Wu, Wei, and Lin]{wu2023scene}
Yang Wu, Pengxu Wei, and Liang Lin.
\newblock Scene graph to image synthesis via knowledge consensus.
\newblock In \emph{AAAI}, pages 2856--2865, 2023.

\bibitem[Xiao et~al.(2018)Xiao, Liu, Zhou, Jiang, and Sun]{xiao2018unified}
Tete Xiao, Yingcheng Liu, Bolei Zhou, Yuning Jiang, and Jian Sun.
\newblock Unified perceptual parsing for scene understanding.
\newblock In \emph{ECCV}, pages 418--434. Springer, 2018.

\bibitem[Xu et~al.(2017)Xu, Zhu, Choy, and Fei-Fei]{xu2017scene}
Danfei Xu, Yuke Zhu, Christopher~B Choy, and Li Fei-Fei.
\newblock Scene graph generation by iterative message passing.
\newblock In \emph{CVPR}, pages 5410--5419, 2017.

\bibitem[Xu et~al.(2022)Xu, Qu, Kuen, Gu, and Liu]{xu2022meta}
Li Xu, Haoxuan Qu, Jason Kuen, Jiuxiang Gu, and Jun Liu.
\newblock Meta spatio-temporal debiasing for video scene graph generation.
\newblock In \emph{ECCV}, pages 374--390. Springer, 2022.

\bibitem[Yan et~al.(2020)Yan, Shen, Jin, Huang, Jiang, Chen, and Hua]{yan2020pcpl}
Shaotian Yan, Chen Shen, Zhongming Jin, Jianqiang Huang, Rongxin Jiang, Yaowu Chen, and Xian-Sheng Hua.
\newblock Pcpl: Predicate-correlation perception learning for unbiased scene graph generation.
\newblock In \emph{ACM MM}, pages 265--273, 2020.

\bibitem[Yang et~al.(2018)Yang, Lu, Lee, Batra, and Parikh]{yang2018graph}
Jianwei Yang, Jiasen Lu, Stefan Lee, Dhruv Batra, and Devi Parikh.
\newblock Graph r-cnn for scene graph generation.
\newblock In \emph{ECCV}, pages 670--685. Springer, 2018.

\bibitem[Yang et~al.(2019)Yang, Tang, Zhang, and Cai]{yang2019auto}
Xu Yang, Kaihua Tang, Hanwang Zhang, and Jianfei Cai.
\newblock Auto-encoding scene graphs for image captioning.
\newblock In \emph{CVPR}, pages 10685--10694, 2019.

\bibitem[Yang et~al.(2022)Yang, Kerce, and Fekri]{yang2022logicdef}
Yuan Yang, James~C Kerce, and Faramarz Fekri.
\newblock Logicdef: An interpretable defense framework against adversarial examples via inductive scene graph reasoning.
\newblock In \emph{AAAI}, pages 8840--8848, 2022.

\bibitem[Ye and Kovashka(2021)]{ye2021linguistic}
Keren Ye and Adriana Kovashka.
\newblock Linguistic structures as weak supervision for visual scene graph generation.
\newblock In \emph{CVPR}, pages 8289--8299, 2021.

\bibitem[Yin et~al.(2019)Yin, Lopes, Shlens, Cubuk, and Gilmer]{yin2019fourier}
Dong Yin, Raphael~Gontijo Lopes, Jonathon Shlens, Ekin~D Cubuk, and Justin Gilmer.
\newblock A fourier perspective on model robustness in computer vision.
\newblock In \emph{NeurIPS}, pages 13276--13286, 2019.

\bibitem[Yoon et~al.(2021)Yoon, Kang, Jeon, Lee, Han, Park, and Kim]{yoon2021image}
Sangwoong Yoon, Woo~Young Kang, Sungwook Jeon, SeongEun Lee, Changjin Han, Jonghun Park, and Eun-Sol Kim.
\newblock Image-to-image retrieval by learning similarity between scene graphs.
\newblock In \emph{AAAI}, pages 10718--10726, 2021.

\bibitem[Yu et~al.(2021)Yu, Chai, Wang, Hu, and Wu]{yu2020cogtree}
Jing Yu, Yuan Chai, Yujing Wang, Yue Hu, and Qi Wu.
\newblock Cogtree: Cognition tree loss for unbiased scene graph generation.
\newblock In \emph{IJCAI}, pages 1274--1280, 2021.

\bibitem[Zareian et~al.(2020{\natexlab{a}})Zareian, Karaman, and Chang]{zareian2020bridging}
Alireza Zareian, Svebor Karaman, and Shih-Fu Chang.
\newblock Bridging knowledge graphs to generate scene graphs.
\newblock In \emph{ECCV}, pages 606--623. Springer, 2020{\natexlab{a}}.

\bibitem[Zareian et~al.(2020{\natexlab{b}})Zareian, Wang, You, and Chang]{zareian2020learning}
Alireza Zareian, Zhecan Wang, Haoxuan You, and Shih-Fu Chang.
\newblock Learning visual commonsense for robust scene graph generation.
\newblock In \emph{ECCV}, pages 642--657. Springer, 2020{\natexlab{b}}.

\bibitem[Zellers et~al.(2018)Zellers, Yatskar, Thomson, and Choi]{zellers2018neural}
Rowan Zellers, Mark Yatskar, Sam Thomson, and Yejin Choi.
\newblock Neural motifs: Scene graph parsing with global context.
\newblock In \emph{CVPR}, pages 5831--5840, 2018.

\bibitem[Zhang et~al.(2019)Zhang, Chao, and Xuan]{zhang2019empirical}
Cheng Zhang, Wei-Lun Chao, and Dong Xuan.
\newblock An empirical study on leveraging scene graphs for visual question answering.
\newblock In \emph{BMVC}, 2019.

\bibitem[Zhang et~al.(2023)Zhang, Stepputtis, Campbell, Sycara, and Xie]{zhang2023robust}
Ce Zhang, Simon Stepputtis, Joseph Campbell, Katia Sycara, and Yaqi Xie.
\newblock Robust hierarchical scene graph generation.
\newblock In \emph{NeurIPS 2023 Workshop: New Frontiers in Graph Learning}, 2023.

\bibitem[Zhang et~al.(2018)Zhang, Cisse, Dauphin, and Lopez-Paz]{zhang2018mixup}
Hongyi Zhang, Moustapha Cisse, Yann~N Dauphin, and David Lopez-Paz.
\newblock mixup: Beyond empirical risk minimization.
\newblock In \emph{ICLR}, 2018.

\bibitem[Zhang et~al.(2022)Zhang, Levine, and Finn]{zhang2022memo}
Marvin Zhang, Sergey Levine, and Chelsea Finn.
\newblock Memo: Test time robustness via adaptation and augmentation.
\newblock In \emph{NeurIPS}, pages 38629--38642, 2022.

\bibitem[Zheng et~al.(2023)Zheng, Lyu, Gao, Dai, and Song]{zheng2023prototype}
Chaofan Zheng, Xinyu Lyu, Lianli Gao, Bo Dai, and Jingkuan Song.
\newblock Prototype-based embedding network for scene graph generation.
\newblock In \emph{CVPR}, pages 22783--22792, 2023.

\bibitem[Zhu et~al.(2021)Zhu, Tremblay, Birchfield, and Zhu]{zhu2021hierarchical}
Yifeng Zhu, Jonathan Tremblay, Stan Birchfield, and Yuke Zhu.
\newblock Hierarchical planning for long-horizon manipulation with geometric and symbolic scene graphs.
\newblock In \emph{ICRA}, pages 6541--6548. IEEE, 2021.

\end{thebibliography}
}

\renewcommand{\thesection}{\Alph{section}}
\renewcommand\thefigure{\Alph{section}\arabic{figure}} 
\renewcommand\thetable{\Alph{section}\arabic{table}}  
\setcounter{section}{0}
\setcounter{figure}{0} 
\setcounter{table}{0} 

\maketitlesupplementary

In this supplementary document, we provide additional details and experimental results to enhance understanding and insights into our proposed HiKER-SGG.
This supplementary document is organized as follows:
\begin{itemize}[leftmargin=0.5cm, itemindent=0cm, itemsep=4pt,topsep=4pt,parsep=0pt]
    \item The hierarchical clustering process and visualization results in Section 3.2 are shown in Section \ref{subsec:hierarchical}.
    \item A more detailed description for both the scene graph and the hierarchical knowledge graph in Section 3.3-3.5 is shown in Section \ref{subsec:graph}.
    \item More discussions about the adaptive refinement in Section 3.7 is in Section \ref{subsec:ar}.
    \item We provide a more comprehensive description of the experimental settings used in our study in Section \ref{subsec:setting}.
    \item A detailed description and visualization of the VG-C dataset are provided in Section \ref{subsec:VG-C}.
    \item We also present more experimental results in Section \ref{subsec:multi} and \ref{sebsec:eff}.
    \item Finally, we discuss the potential limitations of the proposed HiKER-SGG in Section \ref{sec:limit}.
\end{itemize}

\vspace{-6pt}
\section{More Details about HiKER-SGG}
\subsection{Hierarchical Clustering}
\label{subsec:hierarchical}
As we introduced in Section 3.2, we use a hierarchical clustering algorithm to discover both the entity and predicate hierarchies. We provide the pseudocode in Algorithm \ref{alg:hier}. Specifically, hierarchical clustering initializes with individual class names as separate clusters and repeatedly merges the two clusters with the highest similarity until only one cluster remains. During each iteration, it updates the similarity measures of the newly formed cluster with the remaining clusters, ensuring that the most similar clusters are merged at each step. This process results in a hierarchical structure of clusters based on the defined similarity metric.

\begin{figure}[b]
\vskip -20pt
\centering
\includegraphics[width=\linewidth]{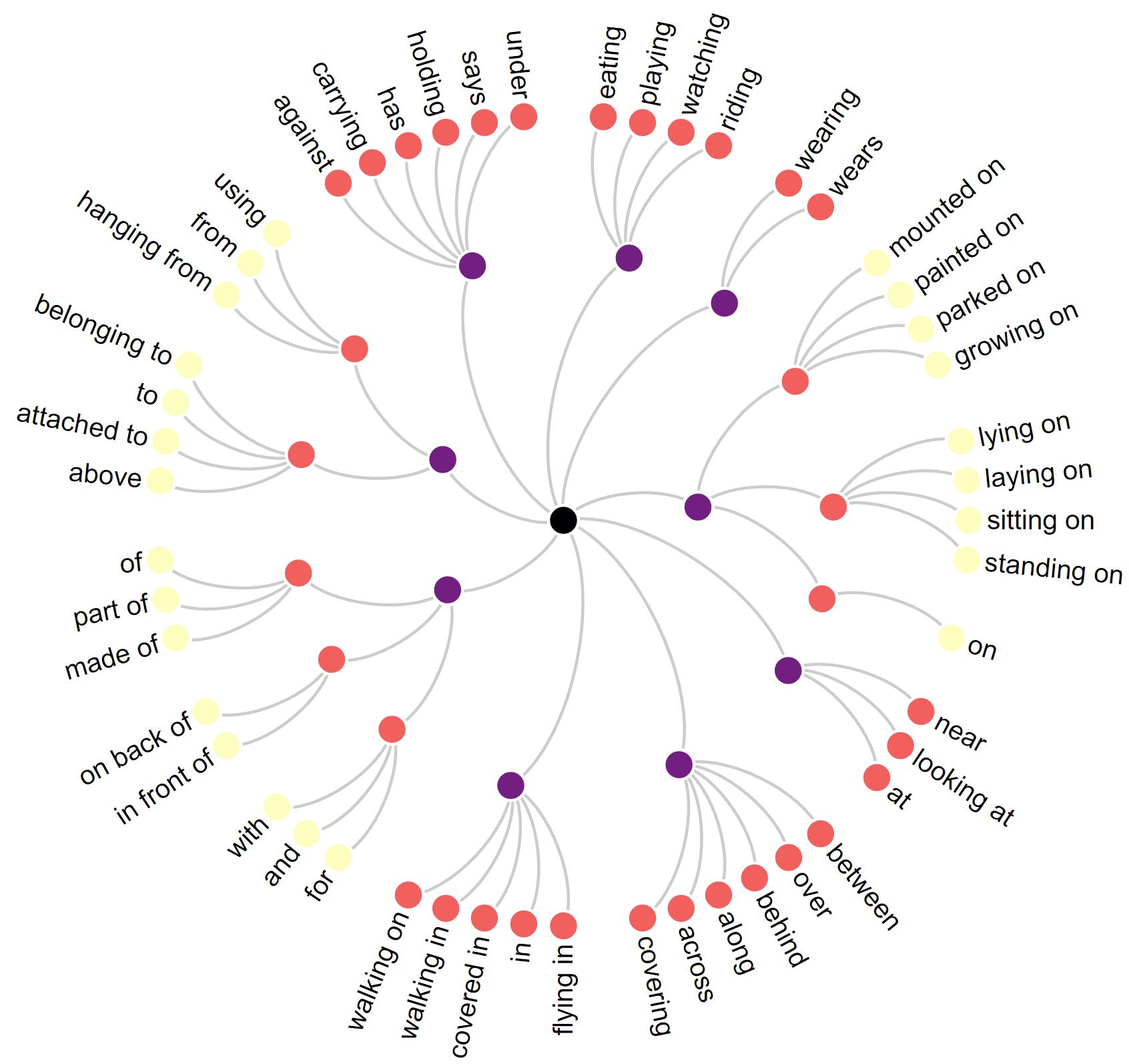}
\includegraphics[width=\linewidth]{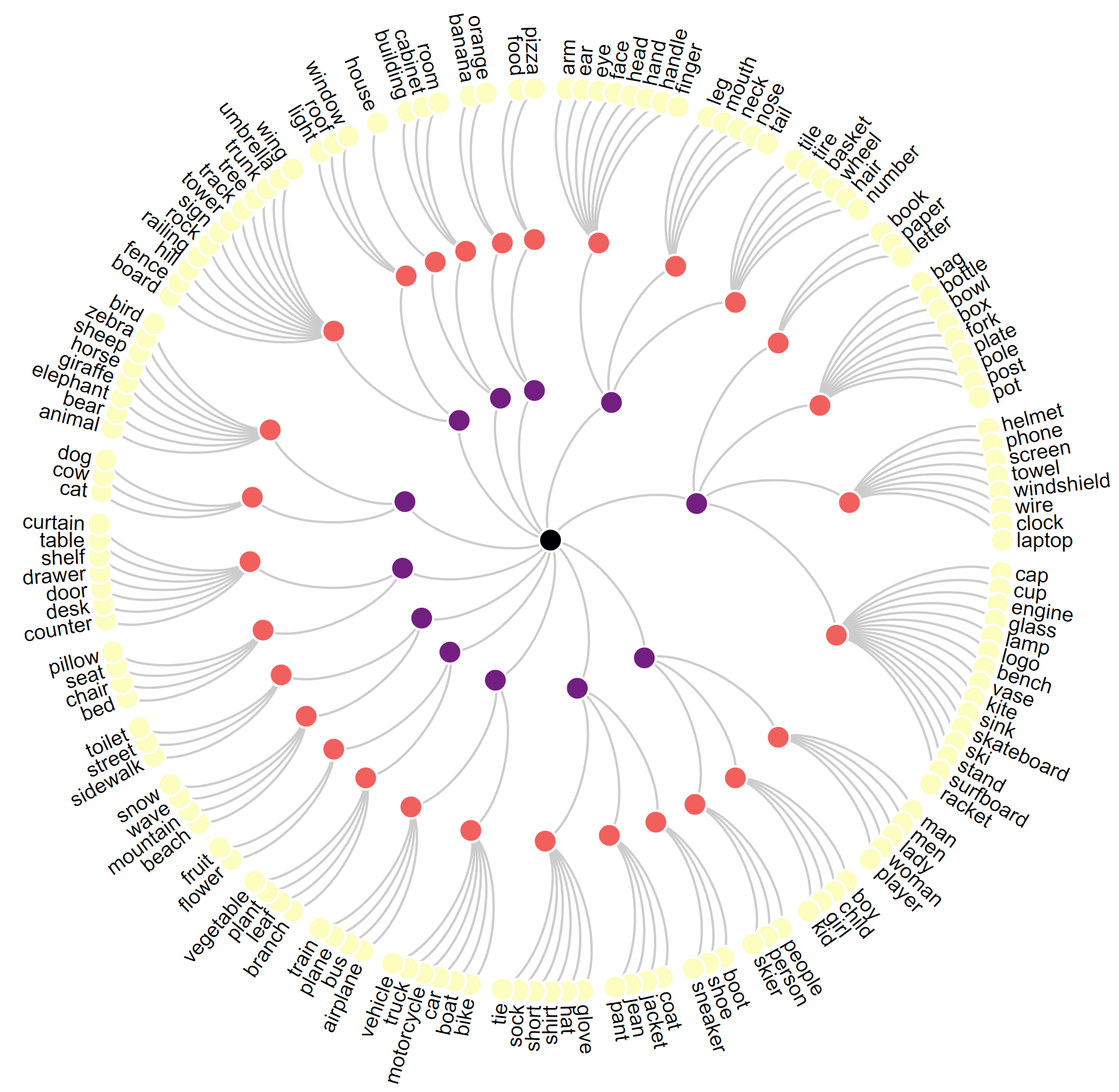}
\vspace{-17pt}
\caption{\textbf{The visualization of the discovered hierarchies on the Visual Genome~\cite{krishna2017visual} dataset}. Top: the discovered hierarchy for 50 predicate classes; Bottom: the discovered hierarchy for 150 entity classes. In this work, we simply utilize 3-level hierarchies for the hierarchical inference process.}
\label{fig:cluster}
\end{figure}

After completing the hierarchical clustering, we select the three lowest-level clusters to conduct our hierarchical inference process defined in Section 3.6. The discovered predicate and entity hierarchies are visualized in Figure \ref{fig:cluster}. Notably, the discovered hierarchical structure reasonably clusters similar classes in the same superclass, for example, (1) \ttt{wearing} and \ttt{wears} in predicate classes, and (2) \ttt{boy}, \ttt{girl}, \ttt{child}, and \ttt{kid} in entity classes. Although most hierarchical relationships are accurately identified, some may appear noisy from a human perspective. Nevertheless, given that our clustering is based on pre-defined similarity metrics, we do not perform additional cleaning and believe our model is equipped to handle these issues. 

In Section 4.3, we have shown that replacing manually configured hierarchical structures with those discovered ones yields a non-trivial 0.4\%$\sim$0.7\% increase in mR@$k$ metrics. These results demonstrate that the hierarchies uncovered by this method, provide more effective guidance for our hierarchical inference approach.

\SetKwRepeat{Do}{do}{while}
\begin{algorithm}[ht]
\begin{normalsize}
\caption{Hierarchical Clustering Algorithm}
\label{alg:hier}
	\LinesNumbered
	\KwIn{Category set $\mathcal{C} = \{\mathcal{C}_i\}_{i=1}^n$;
        Similarity metric $\mathsf{Sim}(\cdot, \cdot)$ defined in Section 3.2.}
        \KwOut{The hierarchy of clusters $L$.}
	Initialize the clusters $L$ with $n$ clusters, each containing a class name\;
	\Repeat{$|L| = 1$}{
	    Find pairs of clusters $\mathcal{C}_1$ and $\mathcal{C}_2$ in $L$ with highest similarity $\mathsf{Sim}(\mathcal{C}_1, \mathcal{C}_2)$\;
        Merge $\mathcal{C}_1$ and $\mathcal{C}_2$ into a new cluster $\mathcal{C}_{12}$\;
        Remove $\mathcal{C}_1$ and $\mathcal{C}_2$ from $L$\;
        \For{each cluster $\Tilde{\mathcal{C}}\in L$}{
            Update the similarity of the created cluster with other clusters with $\mathsf{Sim}(\mathcal{C}_{12}, \Tilde{\mathcal{C}})$\;
        }
        Add this new cluster $\mathcal{C}_{12}$ to $L$\;
        }
\end{normalsize}
\end{algorithm}

\subsection{Scene Graph and Hierarchical Knowledge Graph}
\label{subsec:graph}
\begin{figure}[t]
\centering
\includegraphics[width=\linewidth]{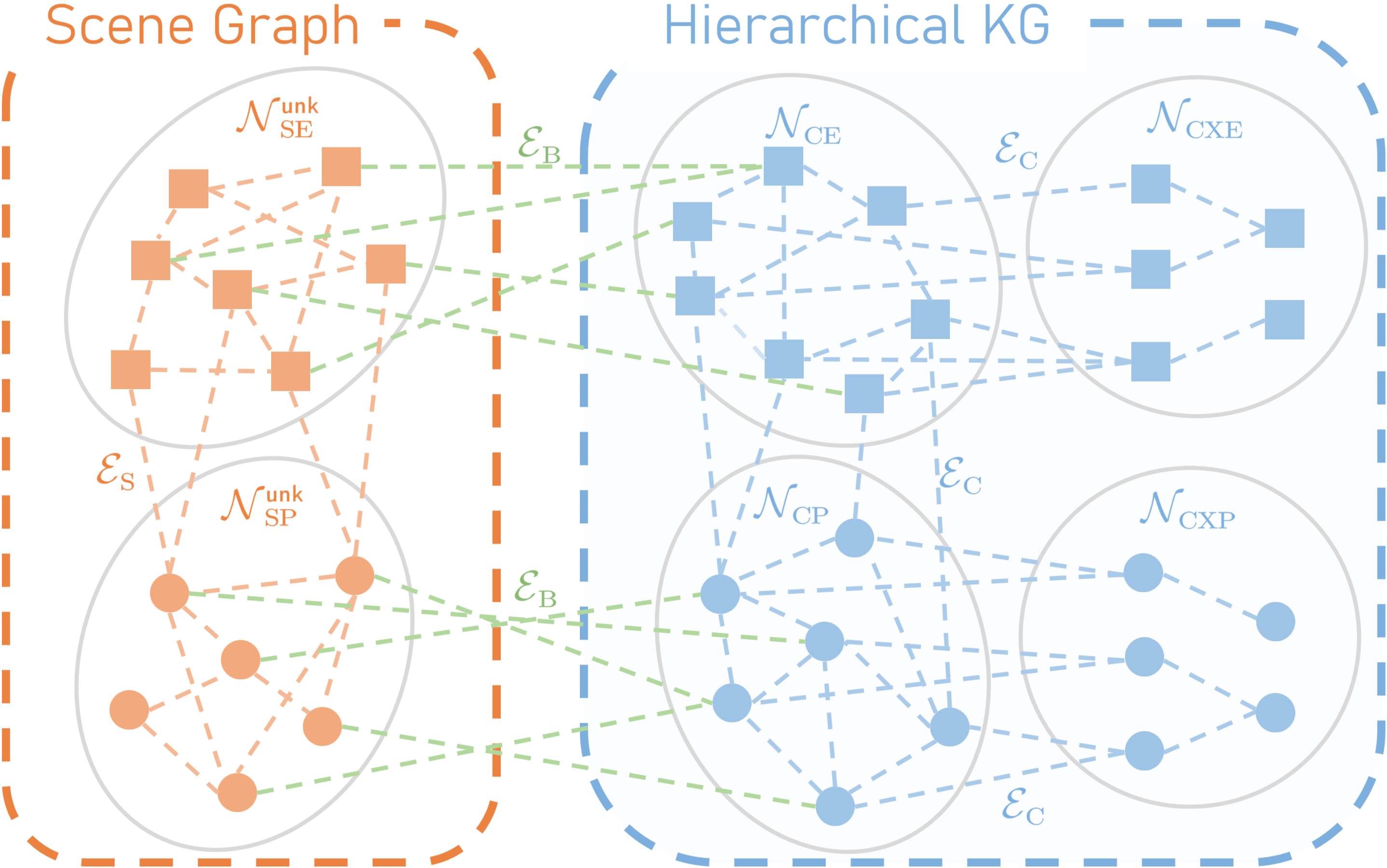}
\vspace{-15pt}
\caption{\textbf{The architecture and notations of our scene graph and hierarchical knowledge graph}. Nodes and edges within the scene graph are \textcolor[RGB]{236,130,64}{orange}, those within the knowledge graph are \textcolor[RGB]{124,175,222}{blue}, and the bridge edges that connect the two graphs are \textcolor[RGB]{137,186,116}{green}.}
\label{fig:graph}
\vspace{-10pt}
\end{figure}

In Figure \ref{fig:graph}, we summarize the architecture and notations of our scene graph and hierarchical knowledge graph we construct in this work. Specifically, we have 6 different types of nodes, as well as 3 types of edges. Below, we detail each one individually.

We have 6 different types of nodes:
\begin{itemize}[leftmargin=0.5cm, itemindent=0cm, itemsep=4pt,topsep=4pt,parsep=2pt]
    \item Commonsense entity node $\mathcal{N}_{\mathrm{CE}}$ in the knowledge graph. We only consider 150 entities from the VG dataset.
    \item Commonsense predicate node $\mathcal{N}_{\mathrm{CP}}$ in the knowledge graph. We only consider 50 predicates from the VG dataset.
    \item Commonsense superclass entity node $\mathcal{N}_{\mathrm{CXE}}$ in the knowledge graph. This includes a set of specialized entity nodes at various levels, corresponding to overarching categories of entities.
    \item Commonsense superclass predicate node $\mathcal{N}_{\mathrm{CXP}}$ in the knowledge graph. This includes a set of specialized predicate nodes at various levels, corresponding to overarching categories of predicates.
    \item Scene entity node $\mathcal{N}_{\mathrm{SE}}$ in the scene graph. Derived from the commonsense entity node, each scene entity (SE) node $\mathcal{N}_\mathrm{SE}$ is additionally linked with a bounding box, \textit{i.e.}, $\mathcal{N}_\mathrm{SE} \subseteq [0,1]^4 \times \mathcal{N}_\mathrm{CE}$.
\item Scene predicate node $\mathcal{N}_{\mathrm{SP}}$ in the scene graph. Originating from the commonsense predicate node, each scene predicate (SP) node $\mathcal{N}\mathrm{SP}$ connects a pair of SE nodes, \textit{i.e.}, $\mathcal{N}_\mathrm{SP} \subseteq \mathcal{N}_\mathrm{SE} \times \mathcal{N}_\mathrm{SE} \times \mathcal{N}_\mathrm{CP}$.
\end{itemize}

We also have 3 types of edges to connect these nodes:
\begin{itemize}[leftmargin=0.5cm, itemindent=0cm, itemsep=4pt,topsep=4pt,parsep=2pt]
    \item Commonsense edges $\mathcal{E}_{\mathrm{C}}$ in the knowledge graph. These edges within the commonsense graph $\mathcal{E}_\mathrm{C}$ delineate relationships between each node pair in both sets, functioning as a reservoir of general knowledge about objects. Examples include connections like \texttt{man}-\texttt{wears}-\texttt{shirt} and \texttt{cat}-\texttt{is}-\texttt{animal}.
\item Scene edges $\mathcal{E}_{\mathrm{S}}$ in the scene graph. These edges encapsulate the relationships within the scene graph, linking scene entities and predicates to denote interactions and spatial relationships in a given scene.
    \item Bridge edges $\mathcal{E}_{\mathrm{B}}$ connecting commonsense nodes and scene nodes. These bi-directional bridge edges link an entity or predicate from the scene graph to its corresponding labels in the commonsense graph. Given the symmetric nature of the relation,  the bridge edges are implemented as bi-directional directed edges with shared weights.
\end{itemize}

\subsection{Adaptive Refinement}
\label{subsec:ar}
\begin{figure}[t]
\centering
\includegraphics[width=\linewidth]{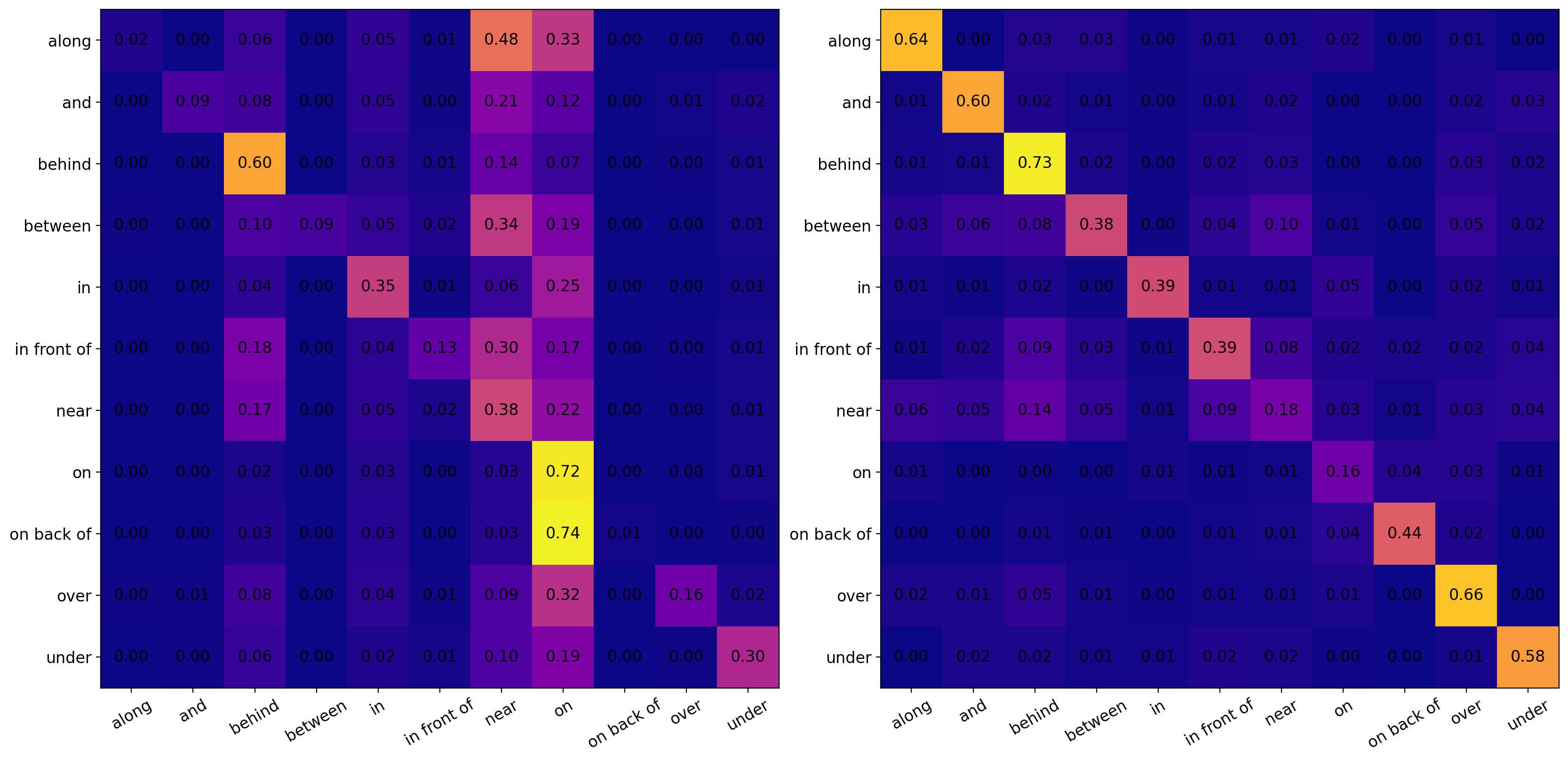}
\vspace{-20pt}
\caption{\textbf{An illustration of adaptive semantic adjustment}. Left: the initial confusion matrix $\mathcal{R}^0$; Right: the confusion matrix $\mathcal{R}^5$ after 5 training epochs.}
\label{fig:confusion}
\vspace{-10pt}
\end{figure}

To provide insights into the adaptive refinement process introduced in Section 3.7, we present a visualization of the initial confusion matrix $\mathcal{R}^0$, along with the updated confusion matrix $\mathcal{R}^5$ after 5 training epochs in Figure \ref{fig:confusion}. The initial confusion matrix $\mathcal{R}^0$ exhibits strong performance in general classes with more samples. In contrast, the updated version  $\mathcal{R}^5$ strives for balanced accuracy across all predicate classes. Simultaneously, rather than the initial matrix which aims to reveal surface-level biases, the updated $\mathcal{R}^5$ shifts its focus to uncovering deeper and subtle correlations between predicate classes, as indicated by more minor values off the diagonal (\textit{e.g.}, \ttt{near} with \ttt{along}/\ttt{and}/\ttt{between}).

\section{More Experimental Results}
\subsection{Experiment Settings}
\label{subsec:setting}

\textbf{Tasks.} Following previous work~\cite{zareian2020bridging,chen2023more}, we assess the effectiveness of our proposed approach in the context of two standard SGG tasks: Predicate Classification (PredCls) and Scene Graph Classification (SGCls). In the PredCls scenario, our model is provided with ground-truth bounding boxes and their associated object classes, with the sole task of predicting the predicate class. In the SGCls scenario, the model is only provided with known bounding boxes while the object classes are treated as unknown, and our SGG model is required to predict both the object and predicate classes.

\looseness=-1
\textbf{Evaluation Metrics.} We evaluate the performance of the SGG models by top-$k$ mean triplet recall (mR@$k$) metric on both the PredCls and SGCls tasks. In specific, mR is the average recall score between the top-$k$ predicted triplets and ground-truth ones across all 50 predicate categories, which promotes unbiased prediction for less frequently occurring predicate classes. A subject-predicate-object triplet is considered a match when all three components are correctly classified, and the subject and object bounding boxes align with an IoU (Intersection over Union) score of at least 0.5. 
In our experiments, we report the mean recall on $k=20,50,100$ to comprehensively evaluate the effectiveness of our method. We also report the constrained (C) and unconstrained (UC) performance results, depending on the presence or absence of the graph constraint. This constraint restricts our SGG model to predict only a single relation between each pair of objects.

\textbf{Implementation Details.} We use the Faster-RCNN~\cite{ren2015faster} as the object detector, which is based on VGG-16~\cite{simonyan2015very} backbone provided by Zellers \textit{et al.}~\cite{zellers2018neural}. 
In our experiments, we train our model for 30 epochs, initializing the learning rate at $1\times 10^{-4}$. 
This learning rate will decrease to $1/10$ of its value after every 10 epochs. 
A single NVIDIA Quadro RTX 6000 GPU is used for all the experiments.

\textbf{Fairness.} To the best of our knowledge, our work is the first to tackle the robustness challenge in SGG, therefore there are  no other established baselines for this task available. However, we do our best to ensure a fair comparison: all models rigorously follow the \textit{\textbf{same}} evaluation protocol stated in Section 4. Our experiments are designed to highlight: (1) Compared to GB/EB-Net, HiKER-SGG enables a more comprehensive and efficient exploitation of KG information. (2) Compared to other methods, the performance gain demonstrates the effectiveness of KG in enhancing SGG.
To ensure a fair comparison with non-graph-based methods, we also conduct an experiment that set the message propagation steps as $t=0$ to isolate the effect of KG. In the PredCls tasks, the mR@50/100 accuracy remains competitive at 34.9\%/37.1\%.

\begin{table}[!t]
  \centering
  \begin{tabular}{c|c}
    \toprule
    Corruption Type& Abbreviation\\
    \midrule
    Gaussian Noise &gaus \\
      Shot Noise &shot\\
     Impulse Noise&imp\\
     Defocus Blur&dfcs \\
     Glass Blur&gls\\
     Motion Blur&mtn  \\
    Zoom Blur&zm \\
    Snow&snw\\
     Frost&frst  \\
     Fog&fg \\
    Brightness&brt\\
    Contrast&cnt\\
     Elastic&els\\
    Pixelate&px\\
    JPEG Compression&jpg\\
    Sunlight glare&sun\\
    Water drop&wtd\\
    Wildfire Smoke&smk\\
    Rain&rain\\
    Dust&dust\\
    \bottomrule
    \end{tabular}
    \caption{\textbf{Abbreviations} of the 20 corruption types in our created corrupted Visual Genome (VG-C) benchmark.}
    \label{tab:abbreviations}
    \vspace{-12pt}
\end{table}

\subsection{Corrupted Visual Genome Benchmark}
\label{subsec:VG-C}
In addition to the clean Visual Genome dataset, we also evaluate our method on the corrupted Visual Genome~\cite{krishna2017visual} (VG-C) dataset, which comprises 20 versions of corrupted images designed to simulate realistic corruptions that may occur in real-world scenarios, thereby providing insights into the models' robustness under various corruption conditions.
Of these corruptions,  the first 15 types of corruption introduced by Hendrycks \textit{et al.}~\cite{hendrycks2018benchmarking} are widely recognized as standard benchmarks for evaluating robustness within the research community. To further align with real-world deployment scenarios, we introduce 5 additional types of \textit{natural} corruptions to our evaluation:
\begin{itemize}[leftmargin=0.5cm, itemindent=0cm, itemsep=4pt,topsep=4pt,parsep=0pt]
    \item \textbf{Sunlight glare}: Sunlight glare refers to the interference caused by excessive sunlight or bright light sources in an image. It typically results in overexposed or washed-out areas in the photo, making it difficult to discern details and colors.
    \item \textbf{Water drop}: Water drop corruption occurs when water droplets or condensation obstruct the camera lens or affect the image sensor. This can create blurry or distorted portions of the image and often results in a hazy or unfocused appearance.
    \item \textbf{Wildfire smoke}: Wildfire smoke corruption pertains to images taken in areas affected by heavy smoke. It causes reduced visibility, a haze or smoky appearance, and can obscure objects in the frame. 
    \item \textbf{Rain}: Rain refers to the presence of falling raindrops in an image. Rain can cause blurriness and distortions, making it difficult to see objects clearly. 
    \item \textbf{Dust}: Dust corruption results from particles or dust settling on the camera lens or sensor. This can lead to the appearance of dark spots or specks in the image, which may obscure details and reduce clarity.
\end{itemize} 

\begin{figure*}[p]
\centering
\includegraphics[width=\textwidth]{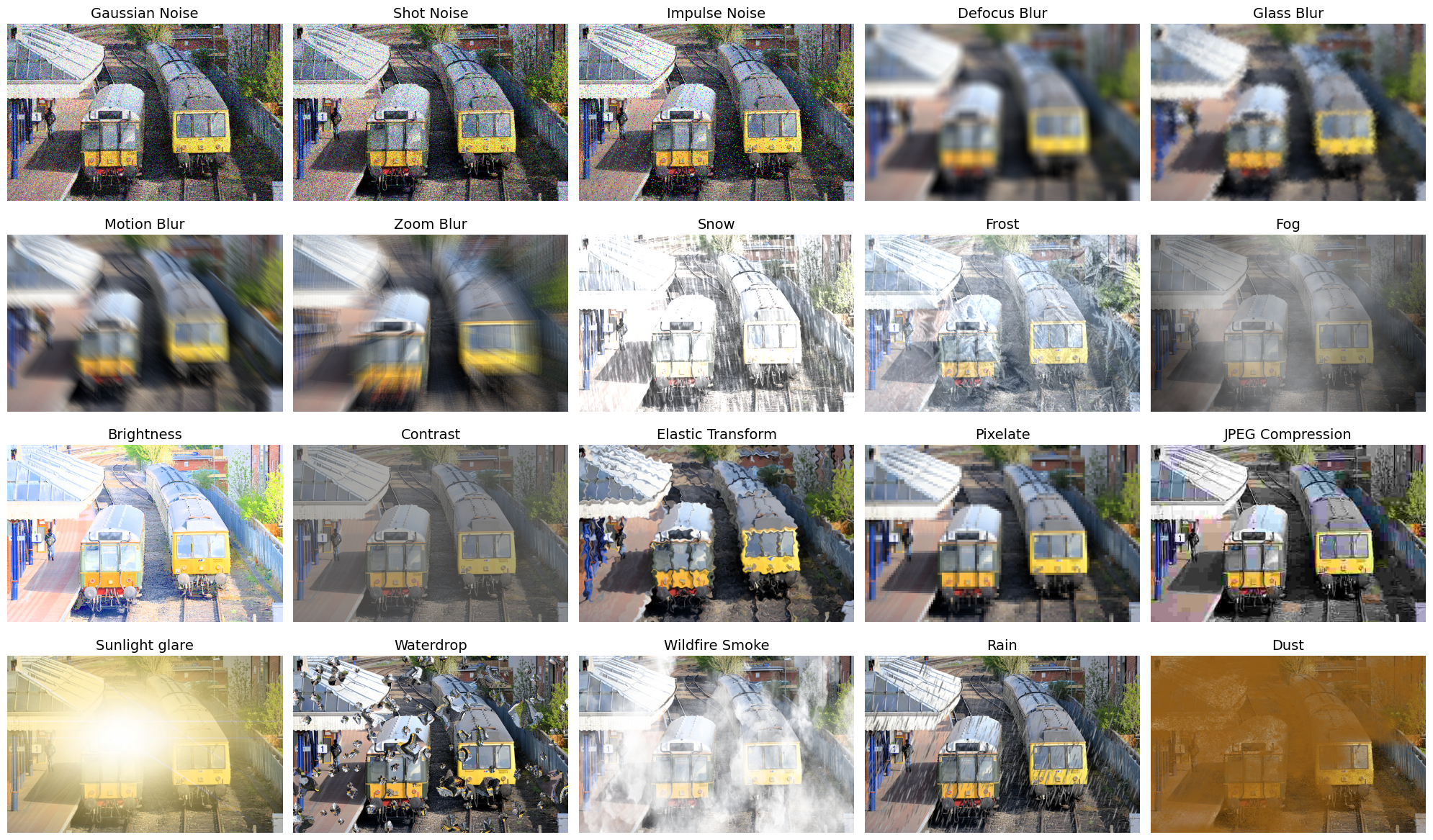}\\ \vspace{15pt}

\includegraphics[width=\textwidth]{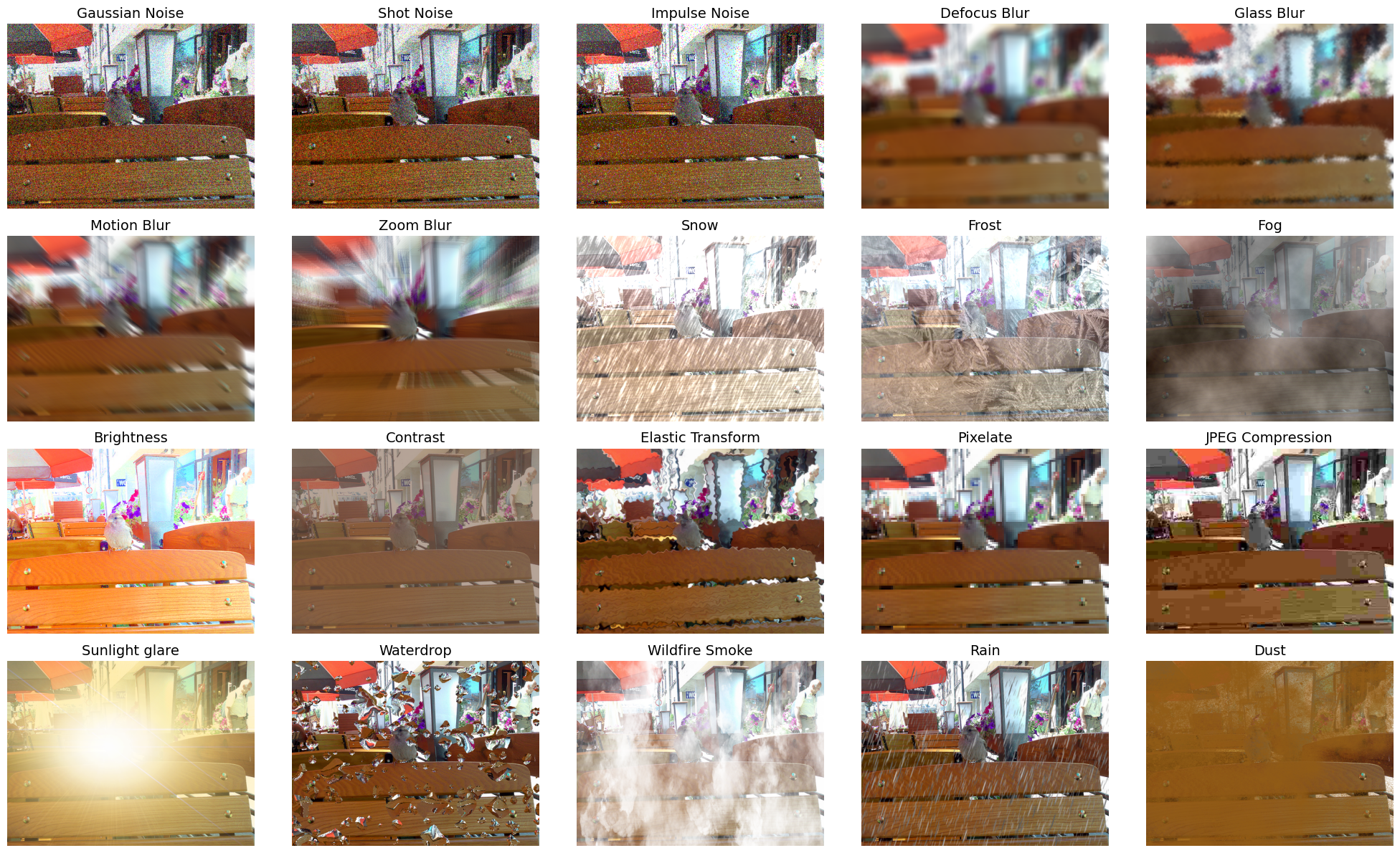}
\vspace{-15pt}
\caption{\textbf{All the 20 corruption types we used in our corrupted experiments}. The first 15 types of corruption are introduced by Hendrycks \textit{et al.}~\cite{hendrycks2018benchmarking}, and we introduce 5 additional types of \textit{natural} corruptions for a more comprehensive and practical evaluation.}
\label{fig:corruption}
\vspace{-7pt}
\end{figure*}

We establish 5 distinct severity levels for each corruption, following Hendrycks \textit{et al.}~\cite{hendrycks2018benchmarking} to facilitate future benchmarking.
Table \ref{tab:abbreviations} presents a summary of the abbreviations used for the various types of corruption. 
To illustrate the effects of these corruptions, we present the corrupted versions of two example images in Figure \ref{fig:corruption}.

We have already made the processing code for generating these corruptions available (VG-C benchmark) at \href{https://github.com/zhangce01/HiKER-SGG}{https://github.com/zhangce01/HiKER-SGG}. This benchmark offers a comprehensive evaluation platform to assess the robustness of SGG models in adverse conditions, and we encourage the formulation of new SGG models to be evaluated using this benchmark, emphasizing the real-world applications of the SGG task.

\begin{table}[t]
\centering
\caption{\textbf{Multi-hop accuracy on the PredCls task using the Visual Genome~\cite{krishna2017visual} dataset}. We compare our method with EB-Net~\cite{chen2023more} method, assessing performance based on both level-1/2 superclass and final subclass accuracy.}
\vspace{-8pt}
\label{tab:multihop}
\resizebox{\linewidth}{!}{
\begin{tabular}{c|c|ccc}
\toprule
&Setting      & mR@20: UC/C & mR@50: UC/C & mR@100: UC/C \\ \midrule
\multirow{3}{*}[-0.3ex]{\centering \rotatebox{90}{EB-Net}} &1-hop     &  51.6 / 50.5 & 68.2 / 63.7 & 79.4 / 68.1
 \\
&2-hop        &   45.4 / 40.2 & 62.8 / 48.9 & 73.7 / 52.0
   \\ 
&\cc 3-hop   &   \cc 39.8 / 30.8 & \cc 54.9 / 36.7 & \cc 66.3 / 39.2
 \\ \midrule
\multirow{3}{*}[-0.3ex]{\centering \rotatebox{90}{Ours}} &1-hop     &  59.6 / 57.8 & 75.6 / 69.1 & 87.7 / 75.3
 \\
&2-hop        &   50.8 / 45.2 & 67.7 / 53.8 & 79.6 / 57.2
   \\ 
&\cc 3-hop   &   \cc 42.1 / 33.4 & \cc 57.9 / 39.3 & \cc 69.2 / 41.2
 \\
\bottomrule
\end{tabular}
}
\vspace{-10pt}
\end{table}

\subsection{Multi-Hop Accuracy}
\label{subsec:multi}
\looseness=-1
To further illustrate the robustness of our method, we compare HiKER-SGG with EB-Net~\cite{chen2023more} by multi-hop mean recall metrics on the Visual Genome dataset in Table \ref{tab:multihop}. Our evaluation criterion is as follows: a 1/2-hop prediction is considered correct if  any of the final predicted predicate classes in the triplets correspond to the true level-1/2 superclass. By designing our model to predict from higher to lower levels, our HiKER-SGG not only achieves state-of-the-art performance in final subclass prediction, but also exhibits superior performance in 1/2-hop superclass prediction, outperforming the baseline method by  an average of 8\% and 5\% in mean recall, respectively.   This performance highlights that when unable to classify to the final subclass, HiKER-SGG tends to more accurately predict the superclass, illustrating the robustness of our hierarchical prediction approach.

\subsection{Inference Time}
\label{sebsec:eff}
In Section 4.3, we have shown that our HiKER-SGG exhibits significantly enhanced robustness with both clean and corrupted images with only about 10\% training costs. In Table \ref{tab:efficiency2}, we also include inference time form comparisons. A single NVIDIA Quadro RTX 6000 GPU is used for all the experiments. When compared to state-of-the-art methods such as GB-Net~\cite{zareian2020bridging} and EB-Net~\cite{chen2023more}, the HiKER-SGG model only extends the inference time by a slight 0.02-0.04 seconds. This minor increase is likely negligible in practical real-world deployment scenarios.

\begin{table}[!t]
\caption{Training time, testing time, and parameter count of HiKER-SGG compared with other methods.}
\vspace{-8pt}
\label{tab:efficiency2}
\resizebox{\linewidth}{!}{
\begin{tabular}{lccc}
\toprule
Method & Training (/epoch) & Inference (/image)     & \# params \\ \midrule
KERN~\cite{chen2019knowledge} & 179.1 min& 0.32 s& 405.2M \\
GB-Net~\cite{zareian2020bridging} & 84.6 min& 0.20 s& 444.6M \\
EB-Net~\cite{chen2023more} & 89.7 min& 0.22 s& 448.8M \\
\cc \textbf{HiKER-SGG} & \cc 101.3 min& \cc  0.24 s& \cc 455.9M \\
\bottomrule
\end{tabular}
}
\vspace{-12pt}
\end{table}

\section{Limitations}
\label{sec:limit}
We identify two potential limitations of our HiKER-SGG method: (1) For each new dataset, a hierarchical structure must be re-discovered, potentially increasing complexity. Additionally, the selection of similarity metrics also includes bias or the prior incorporation by humans. We acknowledge that the choice of measures does reflect a one-time prior human incorporation. However, once determined, the process becomes systematic. This is fundamentally different from the continuous, subjective interventions that characterize the human bias we aim to avoid. 
(2) Our method is tested in corrupted experiments on PredCls and SGCls tasks, assuming the accuracy of detected bounding boxes. However, in cases of severely corrupted images where the object detector fails to recognize objects, our HiKER-SGG method may not perform effectively. However, in our experiments, a simple Faster-RCNN is able to identify nearly 50\% of the GT boxes even under corrupted scenarios; In contrast, given the GT boxes, SGG models can only achieve about 11\% mR@100 in SGCls task. This highlights the practical significance of enhancing the robustness of SGG models.
Besides, we also notice that there is another line of work and benchmarks (\textit{e.g.}, Foggy Cityscapes) focusing on designing robust detectors. Combining our approach with them could further enhance the overall reliability of the system.


{


\end{document}